\documentclass[sigconf]{acmart}

\usepackage{booktabs} 
\usepackage{subcaption}
\usepackage{amsmath}

\setcopyright{none}

\begin{document}
\title{A View Independent Classification Framework for Yoga Postures}

\author{Mustafa Chasmai}
\affiliation{%
  \institution{Computer Science and Engineering}
  \streetaddress{}
  \city{Indian Institute of Technology Delhi}
  \country{}
}
\email{cs1190341@cse.iitd.ac.in}

\author{Nirjhar Das}
\affiliation{%
  \institution{Electrical Engineering}
  \streetaddress{}
  \city{Indian Institute of Technology Delhi}
  \country{}
}
\email{ee3190585@iitd.ac.in}

\author{Aman Bhardwaj}
\affiliation{%
  \institution{School of Information Technology}
  \streetaddress{}
  \city{Indian Institute of Technology Delhi}
  \country{}
}
\email{aman.bhardwaj@cse.iitd.ac.in}

\author{Rahul Garg}
\affiliation{%
  \institution{Computer Science and Engineering}
  \streetaddress{}
  \city{Indian Institute of Technology Delhi}
  \country{}
}
\email{rahulgarg@cse.iitd.ac.in}

\renewcommand{\shortauthors}{}

\begin{abstract}
Yoga is a globally acclaimed and widely recommended practice for a healthy living. Maintaining correct posture while performing a Yogasana is of utmost importance. In this work, we employ transfer learning from Human Pose Estimation models for extracting 136 key-points spread all over the body to train a Random Forest classifier which is used for estimation of the Yogasanas. The results are evaluated on an in-house collected extensive yoga video database of 51 subjects recorded from 4 different camera angles. We propose a 3 step scheme for evaluating the generalizability of a Yoga classifier by testing it on 1) unseen frames, 2) unseen subjects, and 3) unseen camera angles. We argue that for most of the applications, validation accuracies on unseen subjects and unseen camera angles would be most important. We empirically analyze over three public datasets, the advantage of transfer learning and the possibilities of target leakage. We further demonstrate that the classification accuracies critically depend on the cross validation method employed and can often be misleading. To promote further research, we have made key-points dataset and code publicly available.
\end{abstract}

%
%


\maketitle

\section{Introduction}
\label{sec:intro}

\begin{table*}[!htb]
  \caption{Summary of related work in chronological order}
  \label{tab:works}
  \begin{tabular}{ccccccccccc}
    \toprule
    Work &  Asanas & Subjects & Angles &
    \begin{tabular}{c}Key\\Points\end{tabular} & 
    \begin{tabular}{c}Code\\ Open\end{tabular}  & 
    \begin{tabular}{c}Dataset\\ Open\end{tabular}  & 
    Modality \footnote{} & 
    \begin{tabular}{c}Possibility of\\ Target Leakage\end{tabular}  & 
    \begin{tabular}{c}Transfer\\ Learning\end{tabular}\\
    
    \midrule
    
    Islam et al. \cite{islam2017yoga}  & - & 5 & 1 & 20 & No & No & frames & Yes & No\\
    Maddala et al. \cite{maddala2019yoganet}  & 42 & 10 & 1 & 25 & No & No & frames & Yes & No \\
    Yadav et al. \cite{yadav2019real} & 6 & 12 & 1 & 18 & Yes & Yes & frames & Yes & Yes\\
    Verma et al. \cite{verma2020yoga} & \textbf{82} & - & - & - & No & Yes  & images &  No & No \\
    Jain et al. \cite{jain2021three}  & 10 & 27 & 1 & - & No & Yes & videos &  No & No \\
    Gupta et al. \cite{gupta2021yogahelp} & 1 & 20 & 1 & - & No & No & frames &  Yes  & No\\
    \midrule
    Ours  & 20 & \textbf{51} & \textbf{4} & \textbf{136} & Yes & Yes & frames & No & Yes \\
    \bottomrule
  \end{tabular}
  \vspace{2mm}
  \caption*{$^1$ Modality can be interpreted as (1) frames - extract input frames from a video, (2) images - still images not taken from any video, (3) video - use complete video segment as input. Note that frames modality with random splits will lead to target leakage.}
    
  \vspace{-4mm}    
 
\end{table*}

Human Activity Recognition (HAR) is one of the important problems in Computer Vision. It includes accurate identification of the activity being performed by a person or a group with the help of images, videos or raw data collected from Sensors. Wide variety of applications of HAR include but are not limited to active and assisted living (AAL), healthcare monitoring, security and surveillance, tele-immersion, and smart home automation~\cite{ranasinghe2016review}. 

These advances in HAR have been backed by large scale open datasets which have been broadly categorized as action level, behavior level, interaction level and group activities level. A comprehensive category wise list of the datasets along with their details can be referred to in survey~\cite{beddiar2020vision}. 

Yoga is a group of physical and mental practices that originated in ancient India. It aims to bring body, mind, and soul in unison.  Over the past few decades Yoga has gained a tremendous popularity across the globe as an art and science of healthy living. United Nations has declared $21^{st}$ June as \textit{`International day of Yoga'}. Several researchers have studied and evaluated the medical benefits of practicing Yoga ~\cite{bussing2012effects, woodyard2011exploring, ross2010health, wang2009use, gothe2019yoga, groessl2015overview}. Including the recent COVID-19 pandemic times study~\cite{sahni2021yoga} that shows, people who practiced Yoga during the lockdowns experienced lower levels of stress, anxiety, and depression.


\begin{figure}[!htb]
  \centering
  \includegraphics[width=\linewidth]{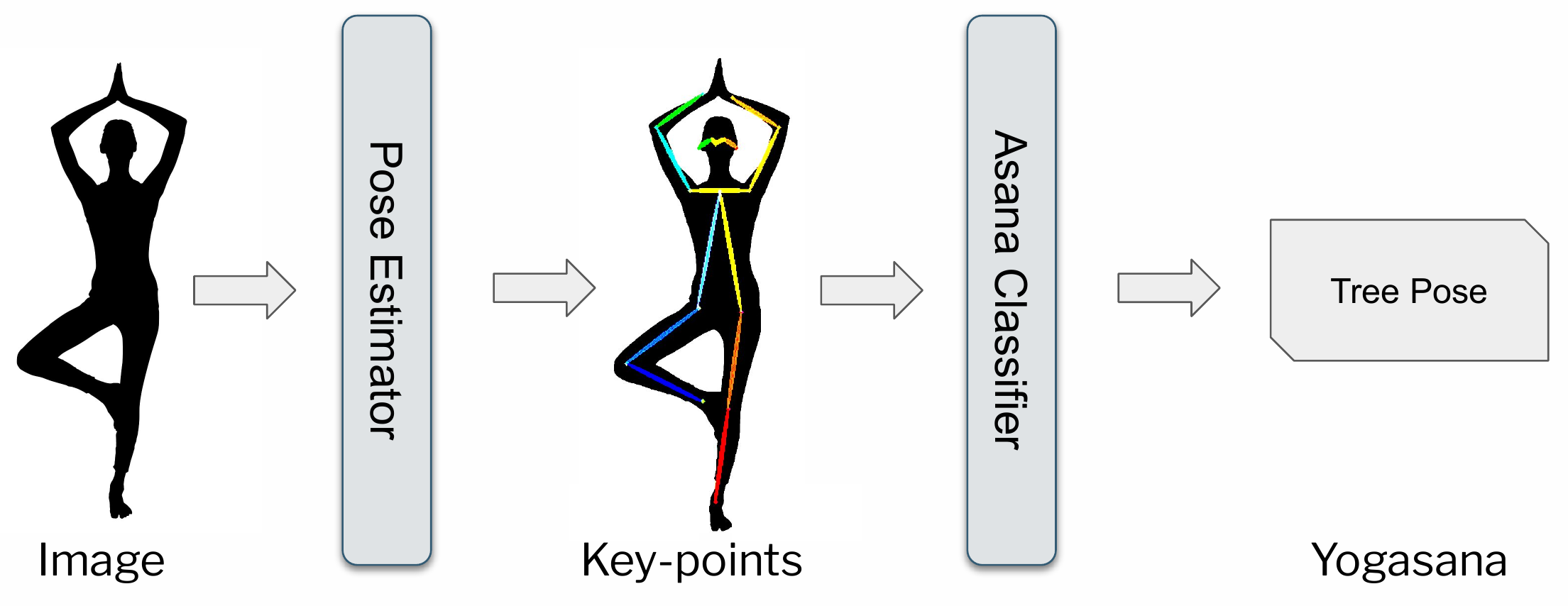}
  \caption{A simplified view of the 2-stage architecture of the Yogasana Classifier with feature extractor and classifier.}
  \label{fig:overall}
  \Description{}
\end{figure}

\vspace{-1mm}

‘Asana’ is the literal Sanskrit translation of posture. To get the maximum benefit from  \textit{Yogasanas} and to prevent any adverse effects and injuries during practice, one need to perform them correctly. This makes the problem of recognition and correction of Yoga pose, an important one. This requires access to a Yoga trainer, which may not be widely available. Therefore, use of ML based Yoga trainer is proposed. However, Yogasanas involve complex body postures which are not seen in general activities. Hence, the existing training datasets available for HAR are likely to fail in accurate estimation of Yoga poses. 

Several studies have created customized dataset for yoga pose estimation~\cite{chen2018computer, chen2014yoga, gupta2021yogahelp, jain2021three, maddala2019yoganet, pandit2020data, trejo2018recognition, verma2020yoga, yadav2019real}. YogaNet~\cite{maddala2019yoganet} is based on 3D joint angular displacement maps and collects 3D information using complex 3D motion capture systems, whereas the rest employ 2D datasets in the form of either videos~\cite{gupta2021yogahelp, jain2021three, trejo2018recognition, yadav2019real} or images~\cite{chen2018computer, chen2014yoga, pandit2020data, verma2020yoga}.

Only a few of the listed datasets have been made publicly available~\cite{verma2020yoga, yadav2019real, jain2021three,pandit2020data}. Yoga-82~\cite{verma2020yoga} is a large scale dataset created from $28.4K$ images of people doing Yogasanas in the wild, collected from various internet sources with the images. These asanas have been classified into 82 poses. Yoga-82 is the most challenging database available till date for Yogasana classification. The other datasets which have been made open are smaller in size, therefore cannot be utilized for Deep Learning model training. For such cases \textbf{Transfer Learning} (TL) proves to be an effective technique and is likely to give better performance. The models using TL are expected to generalize well as compared to the ones that are using limited datasets. 

Human pose estimation is a key attraction in computer vision and has been an active area of research for past couple of decades. Recently, several novel methods have been proposed that can be widely categorized into two types - (i) top-down approach and (ii) bottom-up approach. In the first approach, the architectures first predict a bounding-box around the humans and these bounding boxes are separately processed to predict the joint keypoint coordinates on the human. Prominent works under this approach are AlphaPose \cite{fang2017rmpe}, Simple Baseline \cite{simplebaseline}, HRNet \cite{hrnet}, DARK \cite{dark} and DC-Pose \cite{dcpose}.

In the second approach, the keypoints are first predicted without being assigned to a particular human (among other humans) in the image. Instead, after predicting the keypoints, they are grouped together and assigned to the different humans in the image. Prominent works under this method include OpenPose \cite{cao2019openpose}, MultiPoseNet \cite{multiposenet}, HigherHRNet \cite{higherhrnet}, PifPaf \cite{pifpaf}, SIMPLE \cite{simple}. Since yoga data is limited in size, transfer learning from these models trained on large scale public pose estimation datasets can be employed to improve yoga pose estimation performance.




In this paper we employ a powerful and efficient pose estimation model, AlphaPose \cite{fang2017rmpe}, for pose estimation of the Yoga performer, followed by a simple random forest for classification of the performed yoga. We also experiment with the more recent methods DCPose \cite{dcpose} and KAPAO \cite{kapao} on the Yoga-82 \cite{verma2020yoga} dataset to compare the effect of different pose estimation methods on our overall pipeline. 

Choice of the pose estimation model affects the first stage of the pipeline and good performance here propagates to overall good performance of the pipeline. Our pipeline design allows the choice of pose estimation models and can leverage the developments in this domain.

We also collect an extensive in-house  dataset, a subset of which has been used to evaluate the performance of our model using a three-stage evaluation strategy consisting of evaluation on 1) unseen frames of Yogasanas, 2) unseen subjects, and 3) unseen angles. In summary, the major contributions of this work are three fold:-
\begin{itemize}
    \item An extensive dataset explicitly capturing yoga poses from four camera angles for each of the 51 subjects performing 20 asanas. To promote future research, we make the AlphaPose \cite{fang2017rmpe} inferred body key points dataset and code openly available.
    \item A view independent classification framework, orthogonal to the choice of pose estimation or classification algorithms used, employing a three-stage evaluation strategy to provide a more accurate estimate of generalization capacity of a model.
    \item A simple but effective pipeline involving less computational complexity and real-time inference along with competitive performance exhaustively evaluated on existing datasets.
\end{itemize}

In the rest of the paper we first discuss related work in Section~\ref{section:related-work}. In Section~\ref{section:methodology} we discuss about our data collection method, the two stage architecture for yogasana estimation, possibilities of target leakage and our novel evaluation techniques designed to eliminate target leakage. Experimental results of our and three \footnote{ Images in \cite{pandit2020data} dataset are very similar to those in Yoga-82 \cite{verma2020yoga}, but \cite{pandit2020data} is relatively smaller in size. Thus, we experiment on Yoga-82 \cite{verma2020yoga} and do not explore \cite{pandit2020data}} openly available datasets can be found in Section~\ref{section:results} followed by concluding remarks and future scope in Section~\ref{section:conclusion}. 

\section{Related Work}
\label{section:related-work}

Human pose estimation, being one of the most popular problems in computer vision, has many large scale and openly available benchmark datasets like COCO \cite{lin2014microsoft}, Halpe \cite{fang2017rmpe, li2020pastanet}, MPII \cite{andriluka20142d}, CrowdPose \cite{crowdpose} and HiEve \cite{hive}. Microsoft COCO \cite{lin2014microsoft} dataset is one of the most popular datasets in human pose estimation. It consists of 200,000 images with pose annotation of 17 joint keypoints. MPII \cite{andriluka20142d} dataset contains about 29,000 images of humans performing various activities captured from different angles. This dataset is annotated with 15 body joint keypoints along with their visibility flags. In CrowdPose \cite{crowdpose}, the dataset contains around 20,000 images with about a total of 80,000 humans spanning across all these images. The images of this dataset are sampled from three existing datasets based on a metric called Crowd Index. HiEve \cite{hive} dataset is the largest human pose dataset with a total of {\small$>$}1M poses spanning across 31 videos. This dataset specially focuses on complex and crowded events like entrance/exits at subways, collision, fighting etc. In the absence of a Yoga-pose benchmark, the performance on these datasets can been taken as a proxy for the performance of various pose estimation methods on yoga poses.

Under the top-down approach as described in Section \ref{sec:intro}, the authors of Simple Baseline \cite{simplebaseline} propose a simple but effective pipeline so as to provide a strong baseline for pose estimation methods. Their architecture is based on ResNet \cite{resnet} followed by a few deconvolution layers. With this simple architecture, they achieve 73.7 mAP on COCO dataset and 74.6 mAP and 57.8 MOTA\footnote{mean Average Precision and Multiple Object Tracking Accuracy} score on PoseTrack dataset \cite{posetrack}. In HRNet \cite{hrnet}, the authors propose a deep architecture with parallel high-to-low resolution subnetworks with repeated information exchange across multi-resolution subnetworks. They achieve 75.5 mAP on COCO dataset and 74.9 mAP and 57.9 MOTA on PoseTrack dataset. Somewhat orthogonal to these methods, DARK \cite{dark} investigates the keypoint coordinate representation in human pose estimation architectures. They propose an efficient Taylor-expansion based decoding process of predicted joint heatmap to the keypoint coordinate in the original image space and an  unbiased sub-pixel centred coordinate encoding scheme. With an HRNet-W48 \cite{hrnet} backbone, DARK \cite{dark} achieves 76.2 mAP on COCO dataset. In a different direction, Lite-HRNet \cite{litehrnet} proposes a lightweight High Resolution network that focuses on reducing computational overhead while not degrading the performance significantly. They apply the shuffle block from ShuffleNet \cite{shufflenet} to HRNet \cite{hrnet} to show performance gain. Thereafter, they replace computationally intensive pointwise ($1\times1$) convolutions in shuffle blocks with conditional channel weighting, wherein weights are learnt across all the channels over multiple resolutions. They achieve 69.7 mAP on the COCO dataset and 87.0 PCKh\footnote{Percentage of Correct Keypoints based on the head} on MPII dataset. 

There are several methods that rely on the bottom up approach as it reduces the computational overhead in multi-person pose estimation as the model does not need to process each human in the image separately. OpenPose \cite{cao2019openpose} proposes \textit{part affinity fields} (PAFs) which is a method representing the unstructured pairwise relationship between the body parts of the different humans in the image. They achieve 61.8 mAP on COCO dataset using the vanilla method while the foot+body model achieves 65.3 mAP on the COCO dataset. In MultiPoseNet \cite{multiposenet}, the authors design a deep architecture that consists of a shared backbone of feature extractor which is then fed into two parallel subnets---one is a person detection/segmentation subnet and another is the keypoint detection subnet. The outputs of these two subnets are then fed into a network called Pose Residual Network that assigns the keypoints to the detected persons. This architecture achieves 69.6 mAP on the COCO dataset. HigherHRNet \cite{higherhrnet} tackles the scale variation of humans in the images. It generates high resolution feature pyramid with multi-resolution supervision during training and multi-resolution heatmap aggregation during inference. The pipeline particularly aims at small humans in images and crowded scenes. It achieves 70.5 mAP on COCO dataset. PifPaf \cite{pifpaf} relies on \textit{part intensity fields} (pif) and \textit{part association fields} (paf) to localize body parts and to associate the body parts among each other to fully form the human poses. They also use Laplace loss for regression to encode the uncertainty. This method achieves 66.7 mAP on the COCO dataset. In SIMPLE \cite{simple}, the authors aim to close the gap in performance in terms of accuracy between the top-down approaches and the bottom-up approaches. The pipeline employs mimicking the estimated heatmaps of a high performance top-down approach to transfer the knowledge of high level features from the top-down model to the bottom-up model. The human detection and pose estimation modules share the same backbone and are unified by treating the problems as point learning problems so that both tasks can benefit each other. It achieves 71.1 mAP on the COCO dataset and a 69.5 mAP and a 55.7 MOTA on the PoseTrack dataset. DEKR \cite{dekr} is a simple yet effective pipeline that uses adaptive convolutions and a multi-branch structure for attending to the different pixel regions relevant for different keypoints. The authors argue that to learn keypoints by regression, the model needs to focus on the keypoint regions. This is achieved by the pixel-wise extension of spatial transformer network which activate the pixels near a keypoint allowing the model to learn rich representations from these activated pixels. Further, the multi-branch structure helps the model to focus on the pixels relevant to each keypoint separately, thus learning a disentangled representation. This method achieves 71.0 mAP on the COCO dataset.

Another paradigm in pose estimation is that of regression-based methods, where the keypoint coordinates are directly treated as targets and the model is made to learn a regression-based mapping to the pixel coordinates. Although these methods are computationally less expensive compared to heatmap-based methods, their performances are lower. This is because these models fail to incorporate contextual information which present around the keypoint and also fail to capture the inherent uncertainty in the keypoint annotations, especially in the cases of occlusion and motion blur. A notable method in this area is the Residual Log-likelihood Estimation \cite{rle} that actually achieves performance higher than the SOTA on the COCO dataset. In \cite{rle}, the authors propose a novel and effective regression paradigm with the re-parameterization design and Residual Log-likelihood Estimation (RLE) which, instead of learning the actual distribution of keypoint coordinates, attempts to learn the change of the distribution from the presumed distribution. With an HRNet-w48 backbone and RLE, the authors achieve 75.7 mAP on the COCO dataset. In a similar vein, authors of \cite{kapao} propose a heatmap-free method which they name as KAPAO (\underline{K}eypoints \underline{A}nd \underline{P}oses \underline{A}s \underline{O}bjects), where individual keypoints and sets of related keypoints (poses) are modelled as objects within a dense single-stage anchor-based detection framework. KAPAO solves the problem of single-stage multi-person human pose estimation by simultaneously detecting human pose and keypoint objects and fusing the detections to exploit the strengths of both object representations. KAPAO achieves 70.3 mAP on COCO dataset with the pipeline being 1-2 orders of magnitude faster. We use KAPAO as a pose estimation method in testing on Yoga-82 \cite{verma2020yoga} dataset to observe the effect of various pose estimation models on our pipeline.

AlphaPose \cite{fang2017rmpe} framework consists of three modules---(i) Spatial Symmetric Transformer Network (SSTN) (ii) Parametric Pose Non-Max Suppression (NMP) and (iii) Pose Guided Proposals Generator (PGPG). The SSTN is used extract high-quality single person area in the image from an inaccurate bounding box which can come from a sub-optimal object detector. The Parametric Pose NMS eliminates redundant poses by using a novel pose distance metric. The PGPG is used to augment the training data to generate (sub-optimal) bounding boxes based on the given pose which are used to train the SSTN. It achieves strong results on the MPII \cite{andriluka20142d} benchmark (76.7 mAP) and the COCO dataset (72.3 mAP), and is able to provide a frame rate of 23 fps when fed with video data. Hence we choose AlphaPose \cite{fang2017rmpe} as our pose estimation method.

DCPose \cite{dcpose} aims to solve the problem of multi-person pose estimation in video data. The framework encodes the spatial-temporal keypoint context into localized search scopes, computes pose residuals, and subsequently refines the keypoint heatmap estimations. In particular, the pipeline consists of three task-specific modules---(i) Pose Temporal Merger (PTM) network, which performs keypoint aggregation over three consecutive frames with group convolution, thereby localizing the search range for the keypoint (ii) Pose Residual Fusion (PRF) network, which efficiently obtains the pose residuals between the current frame and adjacent frames and (iii) Pose Correction Network (PCN) comprising five parallel convolution layers with different dilation rates for resampling keypoint heatmaps in the localized search range. This method achieves 79.2 mAP on the PoseTrack dataset. We also experiment with DCPose on Yoga-82 \cite{verma2020yoga} to observe the effect of a different pose estimation method on our pipeline.

Extensive research has been done in the application of pose estimation and classification in Yoga as well. Islam et al. \cite{islam2017yoga} used keypoints to get selected joint angles. They used the deviation in these from a set of reference angles as the asana's accuracy. Although they were not involved in classifying an asana, their experiments and results demonstrated that key-points detected from pose estimation are indeed relevant features for asanas. 

YogaNet \cite{maddala2019yoganet} extended this work by using JADMs \footnote{Joint Angular Displacement Maps} instead of selecting special angles. Using angles instead of key-points allowed them to improve their method's position and size invariance to some extent. However, both this and the previous work relied heavily on key-points detected by Microsoft Kinect \cite{zhang2012microsoft}, which we found to be under-performing compared to more recent pose estimation frameworks like AlphaPose \cite{fang2017rmpe} and OpenPose \cite{cao2019openpose}. This deep learning approach for extracting key points is a relatively inexpensive alternative which only requires RGB images as compared to the depth and infra-red based approach used by Kinect.

Yadav et al. \cite{yadav2019real} used OpenPose \cite{cao2019openpose} for key-point extraction and LSTMs \cite{hochreiter1997long} for exploiting temporal information, building a complete end-to-end pipeline for classification from yoga videos. Instead of using the keypoint co-ordinates, they used intermediate features learned by OpenPose and processed them with CNNs \cite{lecun1998gradient} before putting them into LSTMs \cite{hochreiter1997long}. We improve over this work by using a better performing pose estimator, larger dataset and more concrete evaluations.

Yoga-82\cite{verma2020yoga} is one of the first large scale Yoga classification dataset openly available. It comprises of 28.4K images of people performing one of the 82 specified yogasanas collected from web search engines. A three-level hierarchical structure for labels, with 20 and 6 super-classes on subsequent levels has been provided. However, they formulated their problem as pose \textit{classification} instead of \textit{estimation}, and thus, do not contain key-point annotations. Also, a significant portion of their data consists of clip art images and diagrams, instead of actual human images, which could lead to a poor performance when using TL from methods trained on real-world examples. 

Jain et al. \cite{jain2021three} used a complete end to end architecture for this problem. To utilise the spatio-temporal relationship effectively, they proposed to use 3D CNNs \cite{ji20123d}. However, they have used a small dataset with only 261 videos. 

YogaHelp \cite{gupta2021yogahelp} took a slightly different approach and used various motion sensors to provide feedback and instructions for improvement of a person performing yoga. They extensively explored and tested a single asana for different subjects of varying expertise. They designed different parameters to define the correctness of the asana, and used them for evaluation. They found that using the feedback system, beginners in yoga showed considerable improvement over a small period of four weeks. Their findings demonstrate the overall benefits of this work, and show that a yoga trainer would indeed be very helpful, especially for beginners in the field.

Yadav et al. \cite{yadav2019real} used pre-trained OpenPose \cite{cao2019openpose} model for extracting their key-points. This was the first work which used Transfer Learning for better performance on yoga classification. Using the pre-trained model allowed them to work with a deep learning framework even with a relatively small sized dataset. They used a custom yoga dataset having only 6 asanas, 15 subjects and a uniform camera angle. However, for their frame-by-frame evaluation, they used randomly split data for validation, likely leading to target leakage. This allowed them to obtain 100\% accuracy in three out of the six asanas they considered.

One of the major concerns we found in the existing works was that of \textbf{target leakage} which has been formally defined in section \ref{section:leakage}. In general, target leakage leads to higher accuracy during testing although the underlying model generalization capability may not be comparable to that observed on test set.



In our work, we try to address some of these limitations we found in existing works. We collect Yoga data in a more systematic way. Larger number of subjects and variations across different camera angles allow us to better generalize our models, and obtain more realistic evaluations. Using a three level evaluation strategy, we address target leakage that was prevailing in most existing works. Lastly, we explored a two-stage architecture instead of an end-to-end approach. This allowed us to use Transfer Learning and leverage existing large scale data and extensive research in the form of pre-trained models. In this domain, where large scale datasets are not available, using TL allowed us to improve our performance and paved the way for future works.

\vspace{-1mm}

\section{Methodology}
\label{section:methodology}
Our methodology can broadly be divided into 3 parts. Firstly, we collect an extensive yoga pose dataset while keeping in mind the limitations and shortcomings of existing open datasets. Secondly, we use a 2-stage architecture with human pose estimation for feature extraction and decision tree based classification to get asana prediction for each of the frames given as image input. Finally, we use a tri-level evaluation strategy to evaluate the performance of our model.

\vspace{-1mm}

\subsection{Data Collection}
\label{section:data}

Using Microsoft Kinect \cite{zhang2012microsoft}, we collect data as videos of subjects performing yogasanas. For a better generalization of our models across individuals, 51 volunteers were contacted as subjects for our data. They were each asked to perform 20 yoga poses and 1 still pose (to signify no asana being performed) in a room with stable lighting and proper facilities. Each video lasted about 2-5 minutes. Some asanas were bi-lateral, and the subjects performed these twice, one with each side. The two sides were labelled differently\footnote{Bi-lateral asanas were labelled as \textit{asana\_left} and \textit{asana\_right} separately. Unilateral asanas were labelled normally as \textit{asana}}. For such asanas, the subjects were asked to perform the asana in both directions one after the other, and video timestamps for the start and end time of each direction were recorded as well. 

\begin{figure}[h]
  \centering
  \includegraphics[width=\linewidth]{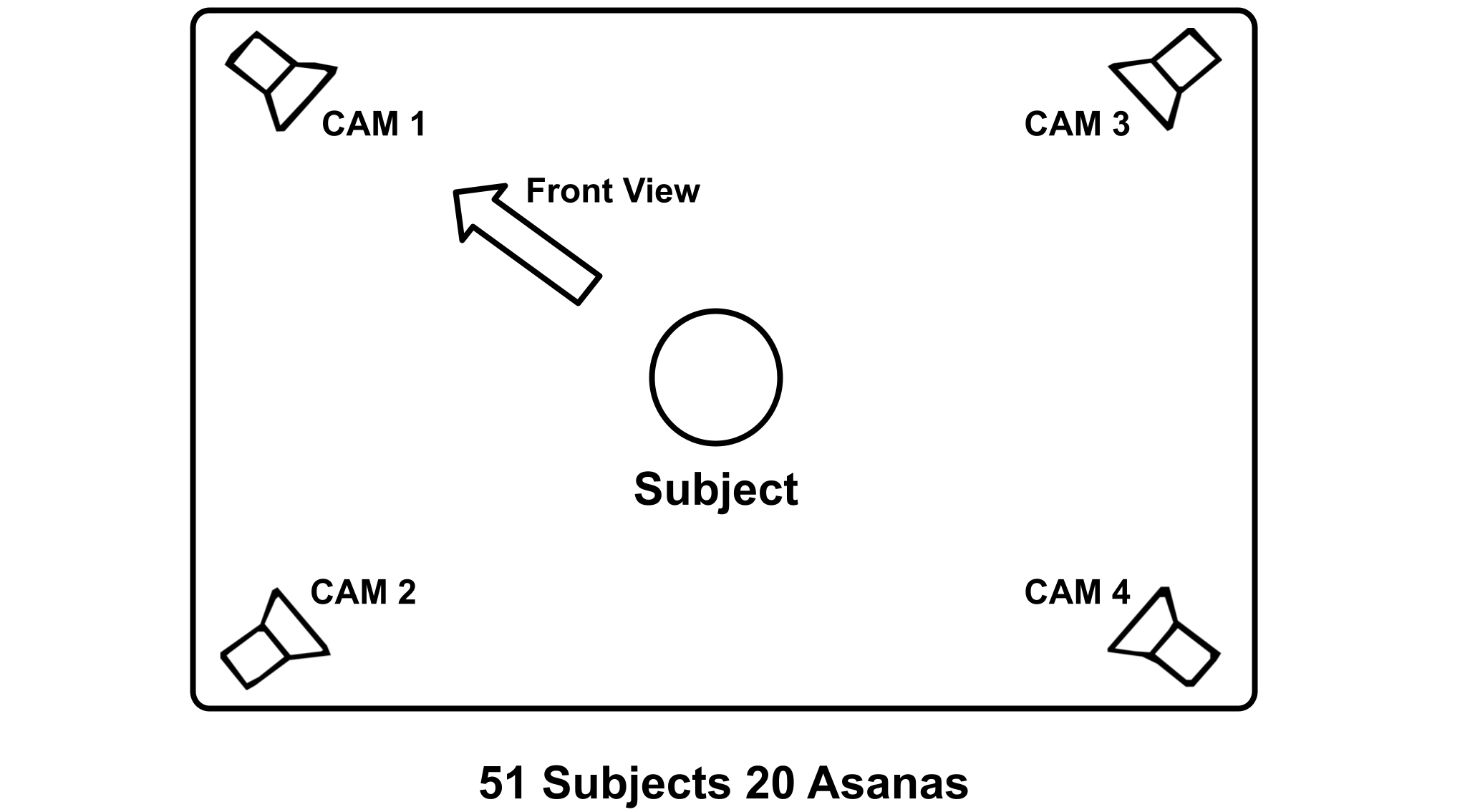}
  \caption{The setup used for data collection}
  \label{fig:setup}
  \Description{}
\end{figure}

Some asanas may be harder to classify from the front facing direction, while they may be easily classified from some other angle. Also, while operating with users in a real-world scenario, the classifier could be fed images from many different orientations. To make our models robust across different view angles, every asana performed by each subject was recorded from 4 different cameras situated in the corners of the room, as shown in Fig~\ref{fig:setup}.

After all the yoga videos were collected, the pose estimator AlphaPose \cite{fang2017rmpe} was first run on all of them together, storing the key-points for each frame in each video. Since we did not have ground truth annotations for the key-points, we used model weights pre-trained on the Halpe Full Body \cite{fang2017rmpe, li2020pastanet} dataset directly for this first stage. Once all the videos were inferred, we started processing it for the second stage of classification. 

For this work, we consider frame-by-frame inference, and so, need frame-wise data for the second classification stage. While collecting the data, we also recorded timestamps corresponding to start and end times of the asana. From each asana, we uniformly sample some frames between the start and end times, such that the frames are equally spaced. We observed that subjects took some time to get into the final pose of the asana. So, we used the frames from the beginning of the video till the start of an asana as `Still' frames, adding an extra class in our classifier. This extra class helped improve the overall performance, since these frames would otherwise be wrongly classified as any one of the asanas. However, we estimate approximately 2-3\% of mislabeled data points where the subject has lost balance while performing the yogasana. The key-points detected by AlphaPose in all the frames and the corresponding asana labels formed the training dataset for our classifier.

In summary, videos of 51 subjects were systematically recorded from 4 different camera angles. From these videos, frames in which actual asana was being performed were uniformly sampled, with a maximum of 200 frames per video segment. A more detailed description of the number of recorded subjects and extracted frames for each yoga pose and camera, as used in the subsequent sections, can be found in the \textcolor{blue}{supplementary material Appendix A}. Katichakrasana was performed by maximum number of subjects, i.e. 47, and different asanas were recorded for different number of subjects. To the best of our knowledge, this is the first yoga dataset explicitly considering asanas from different camera angles, and has the largest number of subjects being recorded. 

\subsection{Pose Estimation}

Pose Estimation is the first and most challenging stage in our model pipeline. A human wanting to distinguish different yogasanas would primarily use the pose and posture of the subject for comparision. Thus, it is only natural to use pose estimation for extracting rich features characterising the pose of the subject. There has already been extensive research in this field and we leverage existing architectures and datasets in our work. Transferring knowledge from openly available large scale datasets allows us to overcome the limitations of scarcity of annotated data to some extent.

AlphaPose \cite{fang2017rmpe} is a popular method for Pose Estimation. It follows a two-step framework where it first detects human bounding boxes and then estimates the pose within each box independently. 
AlphaPose \cite{fang2017rmpe} is comparatively  fast, which makes it ideal for real-time tasks. It is the first open-source system that achieves 70+ mAP (75 mAP) on COCO dataset \cite{lin2014microsoft} and 80+ mAP (82.1 mAP) on MPII \cite{andriluka20142d} dataset. 



\begin{figure}[h]
  \centering
  \includegraphics[height= 50mm] {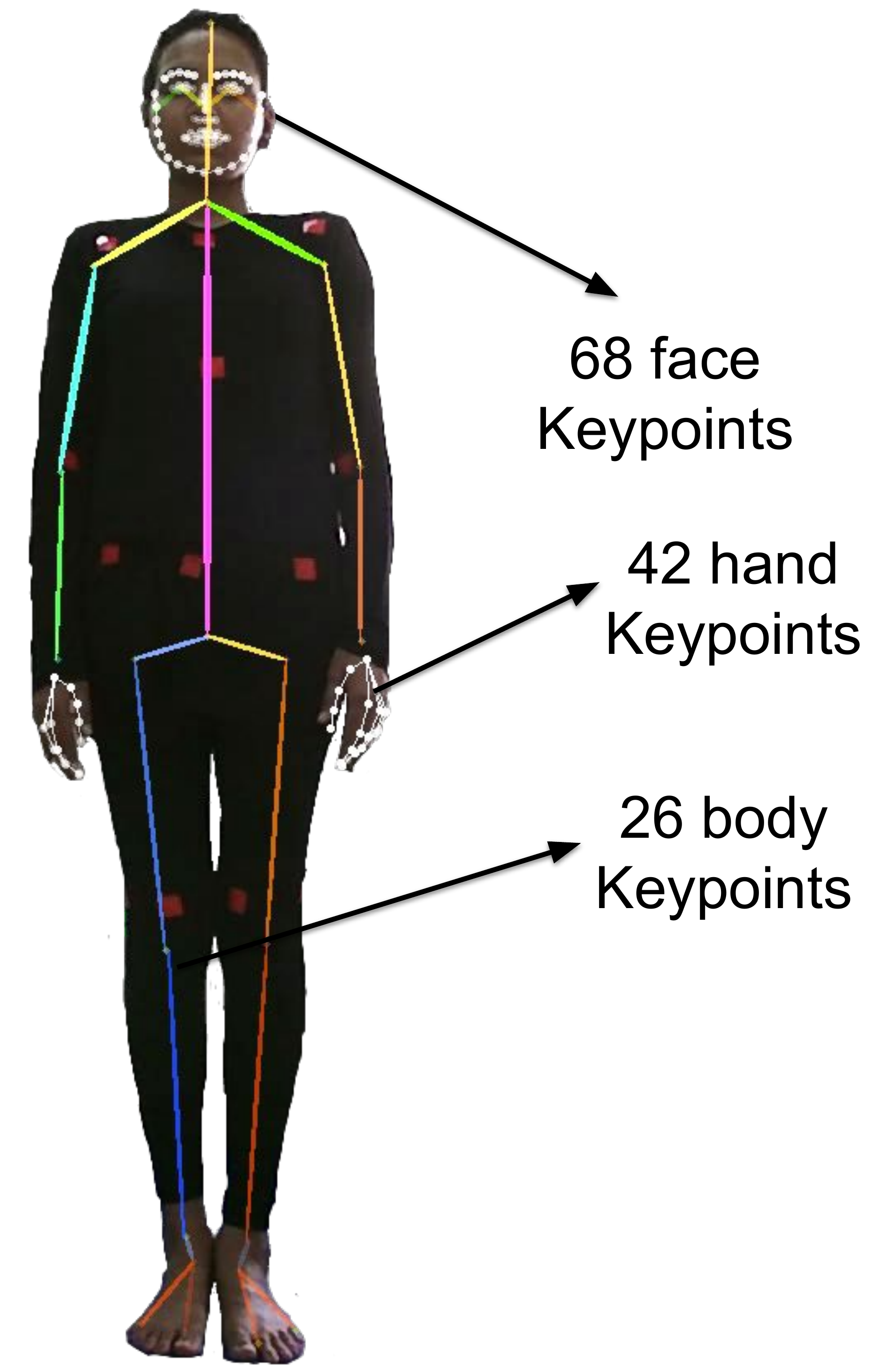}
  \caption{Key-points detected by AlphPose trained on Halpe.}
  \label{fig:key-points}
  \Description{}
\end{figure}

\begin{figure*}[!htb]
  \centering
  \includegraphics[width=\linewidth]{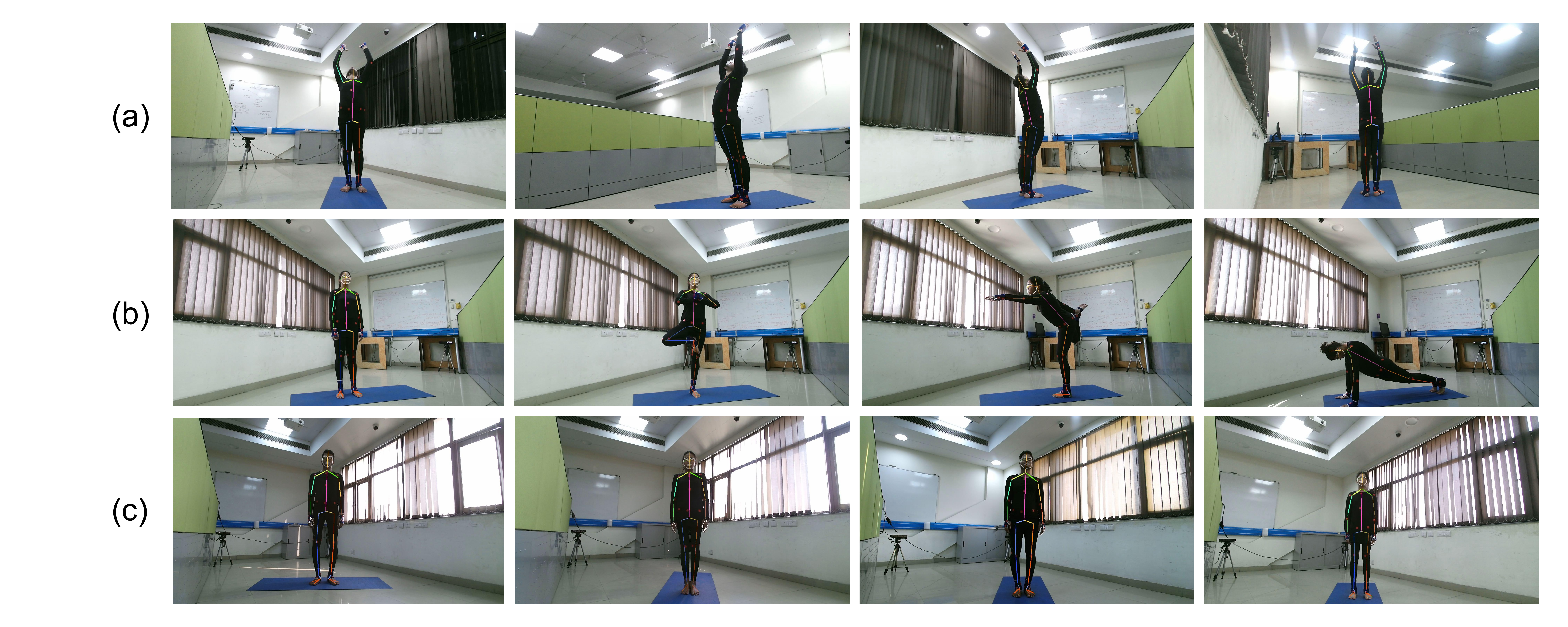}
  \caption{Sample images visualising keypoints detected by AlphaPose. (a) has the same asana and same subject from 4 different camera angles, (b) has the same subject performing different asanas while (c) has different subjects in the `still' pose.}
  \label{fig:alpha_pose_results}
  \Description{}
\end{figure*}


AlphaPose pre-trained on the Halpe Full Body \cite{fang2017rmpe, li2020pastanet} dataset, detects 136 keypoints over a person. As can be seen in Fig~\ref{fig:key-points}, it detects 68 key-points on the person's face, 42 key-points on both hands together and 26 key-points scattered across the rest of the body. Initially, we used the coordinates of all these 136 key-points as our feature vectors. However, since a large number of keypoints in the face and hands are not very relevant for classifying yoga poses, we observed that replacing these sets\footnote{21 left hand key-points, 21 right hand key-points and 68 face key-points are the three key-point sets considered} with the mean, minimum and maximum values of their 10 highest confidence key-points allowed us to obtain better performance. In addition to the keypoints, AlphaPose \cite{fang2017rmpe} also detects bounding boxes around the subject. We found that including the aspect ratio of these bounding boxes as a feature also increased the performance. We also normalise the key-point coordinates with respect to the bounding box, for better size invariance. Thus, we obtain a 71 dimensional (35 key-points x 2 coordinates + aspect ratio) feature vector from our first stage and pass this to the second stage of classification.


\subsection{Classification}

This second stage of the pipeline is a general classification task. The keypoints detected by the pose estimator are used as features by the classifier. To test the validity of our approach, we train a Random Forest \cite{breiman2001random} classifier on the inferred keypoints data. We further explore other boosting methods like ADAboost \cite{schapire2013explaining}, Gradient Boosting \cite{ke2017lightgbm} and Bagging \cite{skurichina1998bagging} classifiers and also an ensemble of the best performing methods. After preparing the dataset of frames and corresponding classes, as explained in Section~\ref{section:data}, we first trained a Random Forest classifier for the second stage of classification. 

\subsection{Evaluation}
\label{section:eval}


\subsubsection{\textbf{Target Leakage}}
\label{section:leakage}
When splitting a dataset randomly into train and test folds, if the data points vary continuously, then the train and test data points will have very little variation between them. Formally, if we split the data in ratio $m:n$, an expected $m$ data points out of any set of $(m+n)$ data points will be in train set, while the rest will be in the test set. Now, if this subset of $(m+n)$ data points are very similar, then the test offers no challenge to the model as it has already memorized the $m$ data points similar to the $n$ data points in test set. Hence, the model is observed to perform very well, sometimes even with $100\%$ accuracy. This is particularly common in time series data, and is formally termed as \textit{Target Leakage.}

In the case of Yoga pose classification, target leakage would be possible if frames from a single video are split into train and test. In most previous works that use video data, the train and test splits are random, and thus target leakage is inevitable. We use video data too, and so, target leakage is a major concern for us. To eliminate the possibility of target leakage, we must ensure that train and test data come from different videos. We experiment with two such methods in addition to frame-wise evaluation as described below.

\subsubsection{\textbf{Three Stage Evaluation}}

First, we consider the normal evaluation strategy of frame-wise testing, where the data is randomly split into train and test sets. 

Second, we consider dividing the data subject-wise. Since each video in our data records only a single subject, this strategy would ensure that the train and test samples are never contiguous and thus, eliminate the possibility of target leakage. In addition to this, it will also allow us to test the generalizability of our models across unseen subjects.

Lastly, we consider dividing the data camera-wise. Again, as each video only has data from one camera, target leakage would be absent if the data is split camera-wise. Further, this strategy would allow us to test the generalizability of our model across unseen camera orientations. The position of the key points would be very different for data captured from different angles, and a good performance across cameras would be a more concrete measure of the model's actual performance for samples in the wild.  

\subsection{Implementation Details}

We use a single RTX 5000 GPU with 16 GB memory for most of our experiments. We use weights trained on Halpe full body \cite{li2020pastanet}, PoseTrack\cite{posetrack} and COCO \cite{lin2014microsoft} for Alphapose, DCPose and KAPAO respectively, with the models detecting 136\footnote{136 keypoints subsequently reduced to 35, as descibed in Methodology}, 17 and 17 keypoints respectively. We used hyperparameters recommended in the corresponding paper themselves and further details regarding their methodology can be found in their work. For our second stage classifier, we use off the self implementations provided by the library sklearn. For  experiments on the Yoga-82 dataset, we use an ensemble of Histogram Gradient Boosting, LightGBM and Random forest, while for the remaining experiments we use a standard Random Forest. We use the Gini criterion with total 500 trees and allow trees to grow until all leaves are pure. Further details on setting up and running the experiments can be found in our code repository.  

\section{Experimental Results}
\label{section:results}

\begin{table}[]
    \caption{Frame-wise results in 10 fold cross-validation. This is equivalent to training \& testing on all 4 cameras in Table~\ref{tab:cam_wise}}
    \label{tab:frame_wise}
\begin{tabular}{clccc}
\toprule
ID & Yoga Pose & Precision & Recall & F1 Score \\
\midrule
0 & Garudasana left & 98.19\% & 99.22\% & 98.70\% \\
1 & Garudasana right & 99.57\% & 99.92\% & 99.74\% \\
2 & Gorakshasana & 99.88\% & 99.98\% & 99.93\% \\
3 & Katichakrasana& 100.00\% & 99.87\% & 99.93\% \\
4 & Natavarasana left & 99.93\% & 99.88\% & 99.91\% \\
5 & Natavarasana right  & 99.75\%  & 99.92\% & 99.83\% \\
6 & Pranamasana left & 99.87\% & 99.83\% & 99.85\% \\
7 & Pranamasana right & 99.83\% & 99.98\% & 99.91\% \\
8 & Tadasana & 99.82\% & 99.88\% & 99.85\% \\
9 & Vrikshasana left & 100.00\% & 99.92\% & 99.96\% \\
10 & Vrikshasana right   & 99.83\%  & 98.32\% & 99.07\% \\
11 & Still & 99.72\% & 99.65\% & 99.68\% \\
\midrule
\multicolumn{2}{c}{Average} & \textbf{99.70\%} &	\textbf{99.70\%} & \textbf{99.70}\%\\
\bottomrule
\end{tabular}
\end{table}

\begin{table}[h]
  \caption{Subject-wise results in 10 fold cross-validation}
  \label{tab:subj_wise}
  \begin{tabular}{clccc}
    \toprule
    ID&Yoga Pose&Precision&Recall&F1 Score\\
    \midrule
    0&Garudasana left& 88.32\%&99.02\%&93.36 \%\\
    1&Garudasana right&96.32\%&99.58\%& 97.93\%\\
    2&Gorakshasana&99.85\%&98.08\%&98.96\%\\
    3&Katichakrasana &99.97\%&99.53\%&99.75\%\\
    4&Natavarasana left&99.97\%&97.47\%&98.70\%\\
    5&Natavarasana right&98.41\%&99.22\%&98.81\%\\
    6&Pranamasana left&99.61\%&94.38\%& 96.93\%\\
    7&Pranamasana right&98.93\%&98.40 \%&98.66\%\\
    8&Tadasana&98.16\%&98.78\%&98.47\%\\
    9&Vrikshasana left&100.0\%&99.87\%&99.93\%\\
    10&Vrikshasana right&99.00\%&93.70\%&96.28\%\\
    11&Still&98.69\%&97.72 \%&98.20\%\\
    \midrule
    \multicolumn{2}{c}{Average}&\textbf{98.10}\%&\textbf{97.97}\%&\textbf{97.99}\%\\
  \bottomrule
\end{tabular}
\end{table}

\begin{table*}[h]
  \caption{Camera-wise results with varying number of cameras trained upon}
  \label{tab:cam_wise}
  \begin{tabular}{cl|ccc|ccc|ccc}
    \toprule
     \multicolumn{2}{c|}{}  &
    \multicolumn{3}{c|}{Training on 3 Cameras} &
    \multicolumn{3}{c|}{Training on 2 Cameras} &
    \multicolumn{3}{c}{Training on 1 Cameras} \\
    ID & Yoga Pose & Precision & Recall & F1 Score & Precision & Recall & F1 Score & Precision & Recall & F1 Score \\
    \midrule
    0 & Garudasana left 
    & 78.48\% & 81.10\% & 79.77\% 
    & 75.88\% & 78.64\% & 77.23\% 
    & 59.22\% & 76.63\% & 64.37\%  \\
    1 & Garudasana right   
    & 83.46\% & 87.79\% & 85.53\% 
    & 71.04\% & 82.88\% & 76.28\%
    & 67.18\% & 79.78\% & 72.36\%  \\
    2 & Gorakshasana    
    & 77.97\% & 68.84\% & 72.62\% 
    & 73.86\% & 77.03\% & 75.30\% 
    & 74.59\% & 57.68\% & 64.90\% \\
    3 & Katichakrasana  
    & 80.22\% & 71.15\% & 75.39\% 
    & 61.16\% & 77.56\% & 67.11\%
    & 73.18\% & 64.48\% & 64.45\%  \\
    4 & Natavarasana left   
    & 99.94\% & 99.48\% & 99.71\% 
    & 99.66\% & 99.47\% & 99.57\%  
    & 96.80\% & 99.42\% & 98.05\% \\
    5 & Natavarasana right  
    & 68.86\% & 47.00\% & 55.18\% 
    & 64.49\% & 46.01\% & 53.25\% 
    & 54.65\% & 37.16\% & 43.73\% \\
    6 & Pranamasana left   
    & 60.10\% & 82.26\% & 69.36\% 
    & 52.30\% & 63.28\% & 57.15\% 
    & 37.31\% & 50.98\% & 42.95\% \\
    7 & Pranamasana right  
    & 77.76\% & 89.18\% & 83.05\% 
    & 74.94\% & 73.57\% & 74.03\%
    & 55.02\% & 55.78\% & 55.29\% \\
    8 & Tadasana   
    & 98.07\% & 96.51\% & 97.28\% 
    & 96.67\% & 90.37\% & 93.34\%
    & 78.39\% & 85.73\% & 79.82\% \\
    9 & Vrikshasana left   
    & 95.02\% & 93.03\% & 93.98\%  
    & 96.16\% & 90.29\% & 93.08\%
    & 94.26\% & 78.43\% & 84.61\% \\
    10 & Vrikshasana right   
    & 76.66\% & 74.42\% & 75.44\% 
    & 69.92\% & 59.92\% & 64.17\%
    & 68.84\% & 44.72\% & 51.14\% \\
    11 & Still    
    & 86.70\% & 85.66\% & 85.98\% 
    & 92.39\% & 70.50\% & 78.95\%
    & 76.08\% & 65.67\% & 69.53\% \\
  \midrule
    \multicolumn{2}{c|}{Average} & \textbf{81.94\%} & \textbf{81.37\%} & \textbf{81.11\%} & 77.37\% & 75.79\% & 75.79\% & 69.63\% & 66.37\% & 65.93\%\\
  \bottomrule
\end{tabular}
\end{table*}

Although we recorded data for 20 asanas, we only use a subset of these for our evaluation. We use 72k extracted frames spanning 11 asanas and a still class, giving a total of 12 classes. 
These 12 classes include left and right asanas for bilateral ones. Each asana has a total of 6000 samples, leading to a highly balanced dataset. For still class, we take all the frames in a video before the start of the actual asana, with a buffer of 1 second. We realise that this buffer may not be sufficient, and as a result, a part of our data may be mislabelled.
More details about the size of our dataset and the camera and subject wise distributions can be found in the \textcolor{blue}{supplementary material Appendix A}.

\subsection{Frame-wise Evaluation}
\label{section:frame_wise}

Frame-wise evaluation is the standard evaluation strategy used by most of the existing works. We evaluate our model using vanilla 10-fold cross validation. The class-wise results can be seen in Table~\ref{tab:frame_wise}. The random forest classifier obtained mean precision, recall and F1 scores of 99.70\% each. Such a high performance due to the explained problem of target leakage (Section~\ref{section:leakage}) can be misleading and can also be witnessed in other studies reporting near perfect accuracy for their classifier.

\subsection{Subject-wise Evaluation}

Our dataset has 51 subjects performing different yoga poses. The subjects are of varying build, ages and genders. Irrespective of the characteristics of the subject, the pose they take during an asana should be similar. Thus, the classifier should be able to generalize across subjects. To test this, we create folds in which the dataset is split subject-wise into a 9:1 ratio. After training our classifier on the first set of subjects, we test on the remaining. These test subjects would be unseen by the model, and thus, this strategy prevents the model to simply memorise the train subjects, and eliminates the possibility of target leakage (Section~\ref{section:leakage}).

We generate 10 such folds, with randomly selected subjects, in rotation, such that each subject gets tested exactly once. The class wise results of these experiments can be found in Table~\ref{tab:subj_wise}. As can be seen, the random forest performs well even across subjects, with an average precision and recall of 98.10\% and 97.97\%, leading to an F1 score of 97.99\%. Compared to the frame wise results, these results are consistently around 1-2\% lower for all three metrics we use for evaluation.



\vspace{-1mm}

\subsection{Camera-wise Evaluation}

We expect camera-wise evaluation to be more challenging than the previous two methods. Firstly, many asanas can be very difficult to identify from side or back views. Secondly, the same asana, viewed from different cameras would have very different key-point co-ordinates, and generalizing to such wide range of variations is challenging. Lastly, some angles would have much higher occlusion than others, leading to poorer pose estimation performance as well. The poor performance in the first stage is propagated to the second stage classifier as well. 

The results can be seen in Table~\ref{tab:cam_wise}. The results are consistently lower than those obtained from earlier methods, with average precision and recalls of only 81.94\% and 81.37\% respectively in the case training on 3 camera angles. We also observe that changine the number of angles being used in training directly impacts the performance. A clear decreasing trend can be seen in the performances of training on 3, 2 and 1 cameras in Table~\ref{tab:cam_wise}. This pattern suggests that including a wider variety of camera angles while training tends to give better results. Including all four camera angles would be equivalent to the earlier two evaluation strategies, both of which gave significantly better results.

We perform the same tri-level testing with key-points detected by Microsoft Kinect \cite{zhang2012microsoft} as well. Using these key-points, the mean F1 scores for frame-wise, subject-wise and camera-wise were 69.49\%, 57.91\% and 35.21\% respectively. These are considerably lower than the corresponding scores we obtained using AlphaPose \cite{fang2017rmpe} key-points, with the same second stage classifier. Thus, a good pose estimator is essential for good performance in our method.

\subsection{Experiments on Yadav et al. dataset}

Yadav et al. \cite{yadav2019real} collected an in-house dataset consisting of 88 videos, with 15 subjects being recorded across 6 asanas. We suspect that their frame-wise evaluation may have target leakage, and evaluate our model on their data both frame-wise and subject-wise.

Using all the frames extracted from the videos together, the authors were able to achieve 99.04\% accuracy on their test set. As can be seen in Table~\ref{tab:yadav_frame}, we obtain similar results while training on only 200 uniformly extracted frames from each video. However, going through the extracted frames, we observed that some frames were having the subject in \textit{transition}, and thus would most likely be mislabelled. We estimate that around 3.5\% of our extracted frames were mislabelled. With this in mind, an accuracy greater than 96.5\% clearly indicates that the model is over-fitting, and that there may be target leakage.

\begin{table}[]
    \caption{Frame-wise results for Yadav et al. \cite{yadav2019real} dataset}
    \label{tab:yadav_frame}
\begin{tabular}{clccc}
\toprule
ID & Yoga Pose & Precision & Recall & F1 Score \\
\midrule
0 & Bhujangasana & 98.85\% & 99.44\% & 99.14\%\\
1 & Padamasana & 98.38\% & 99.79\% & 99.08\% \\
2 & Shavasana & 99.59\% & 98.29\% & 98.93\% \\
3 & Tadasana & 99.87\% & 99.13\% & 99.50\% \\
4 & Trikonasana & 99.81\% & 99.58\% & 99.69\%\\
5 & Vrikshasana & 99.37\% & 99.40\% & 99.38\%\\
\midrule
\multicolumn{2}{l}{Average (Ours)} & \textbf{99.31\%} &	\textbf{99.27\%} & \textbf{99.29\%}\\
\multicolumn{2}{l}{Average (Yadav et al. \cite{yadav2019real})} & 98.97\% &	99.11\% & 99.03\%\\

\bottomrule
\end{tabular}
\end{table}

\begin{table}[]
    \caption{Subject-wise results for Yadav et al. \cite{yadav2019real} dataset}
    \label{tab:yadav_subject}
\begin{tabular}{clccc}
\toprule
ID & Yoga Pose & Precision & Recall & F1 Score \\
\midrule
0 & Bhujangasana & 94.41\% & 95.83\% & 95.12\%\\
1 & Padamasana & 92.33\% & 94.96\% & 93.63\% \\
2 & Shavasana &  98.73\% & 95.51\% & 97.09\% \\
3 & Tadasana & 92.24\% & 94.30\% & 93.26\%\\
4 & Trikonasana & 98.73\% & 98.80\% & 98.77\%\\
5 & Vrikshasana & 94.76\% & 91.07\% & 92.88\%\\
\midrule
\multicolumn{2}{c}{Average} & \textbf{95.20\%} &	\textbf{95.07\%} & \textbf{95.12\%}\\
\bottomrule
\end{tabular}
\end{table}

\begin{figure*}
    \centering
    \begin{subfigure}[b]{0.48\textwidth}
         \centering
         \includegraphics[width=\textwidth]{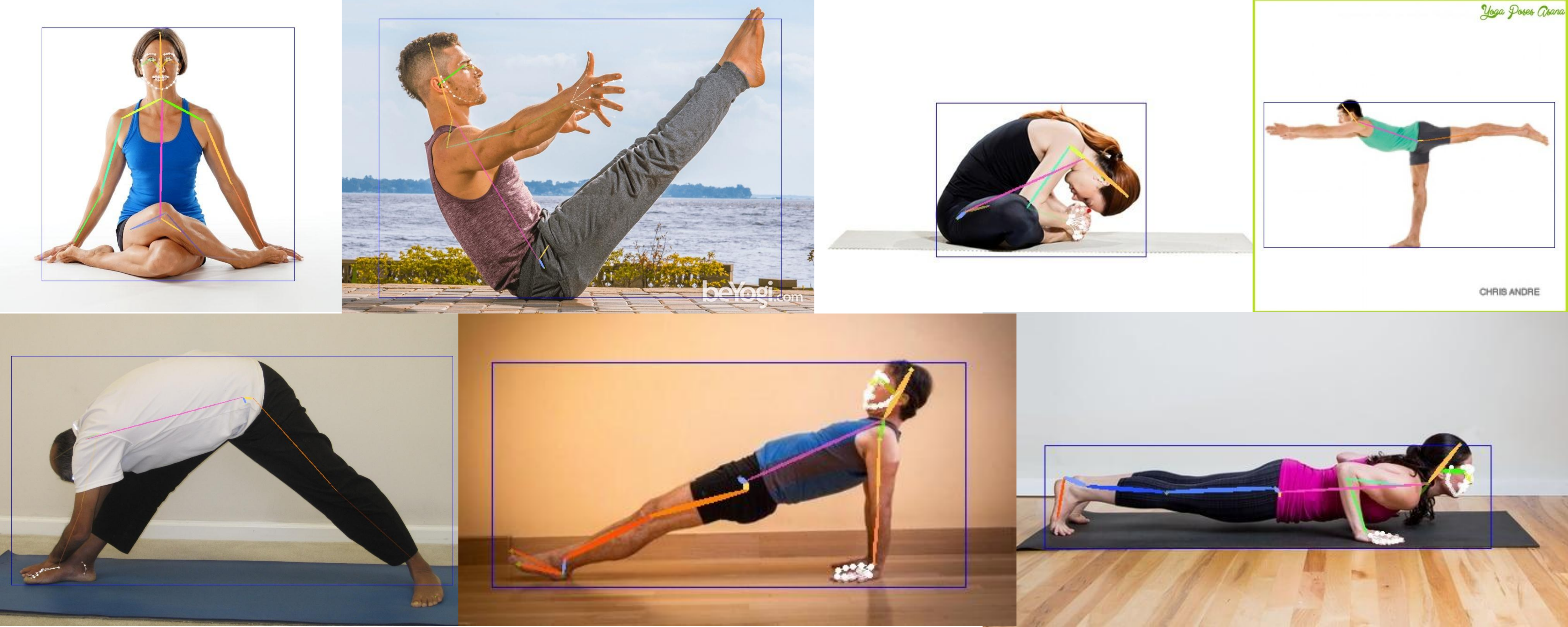}
         \caption{Some example images from Yoga-82 that were mis-classified. Top leftmost is an example of inversion of legs. Top rightmost is a very low resolution image, leading to poorer predictions. Rest of the examples have poor predictions due to occluded body parts.}
     \end{subfigure}
    \hfill
     \begin{subfigure}[b]{0.24\textwidth}
         \centering
         \includegraphics[width=\textwidth]{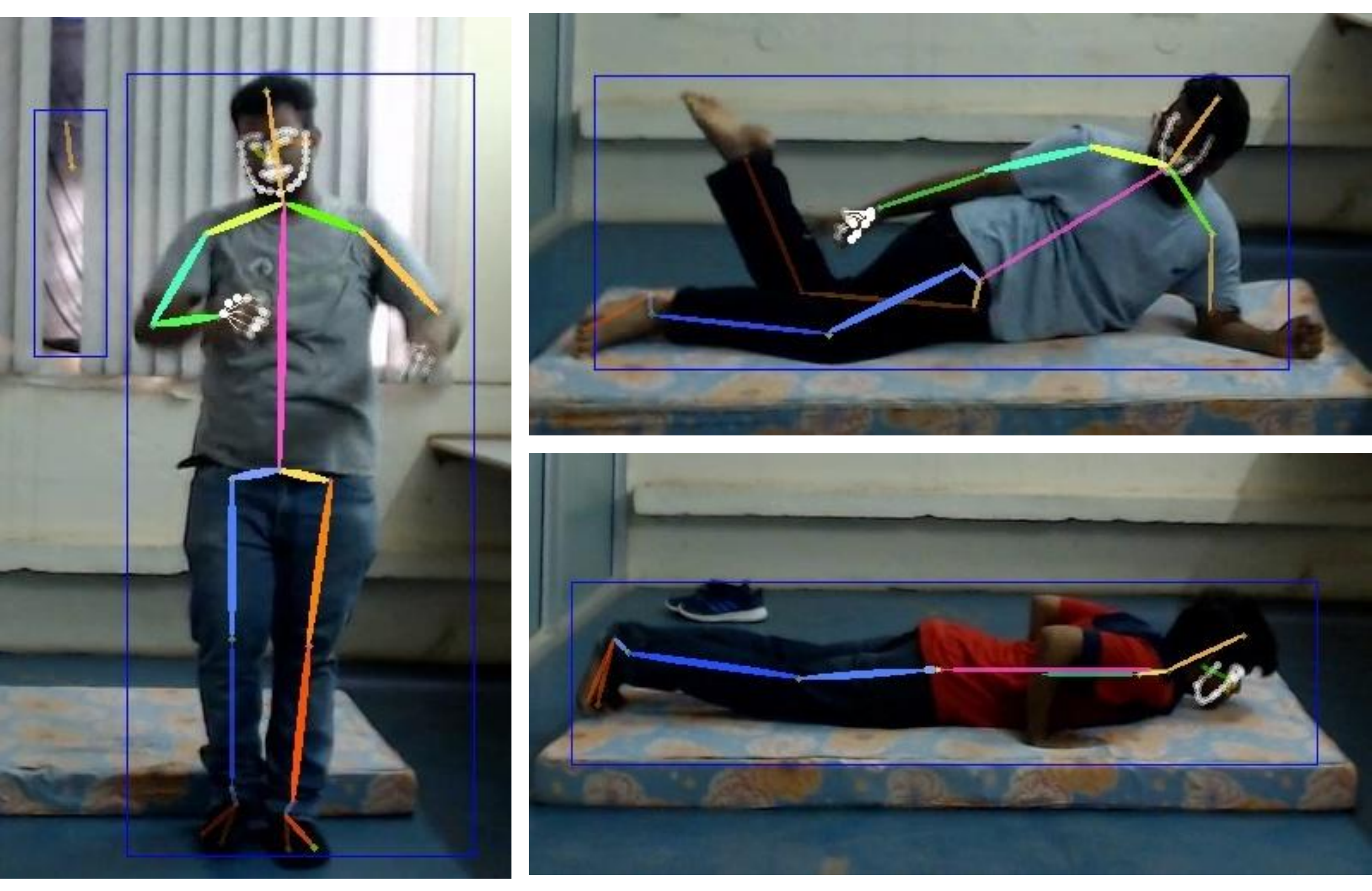}
         \caption{Mislabelled transition frames. The subject’s pose does not match the labelled asana. In left, the subject has not started pose yet. In top right, he is changing, while in botttom right, he has completed.}
     \end{subfigure}
     \hfill
     \begin{subfigure}[b]{0.24\textwidth}
         \centering
         \includegraphics[width=\textwidth]{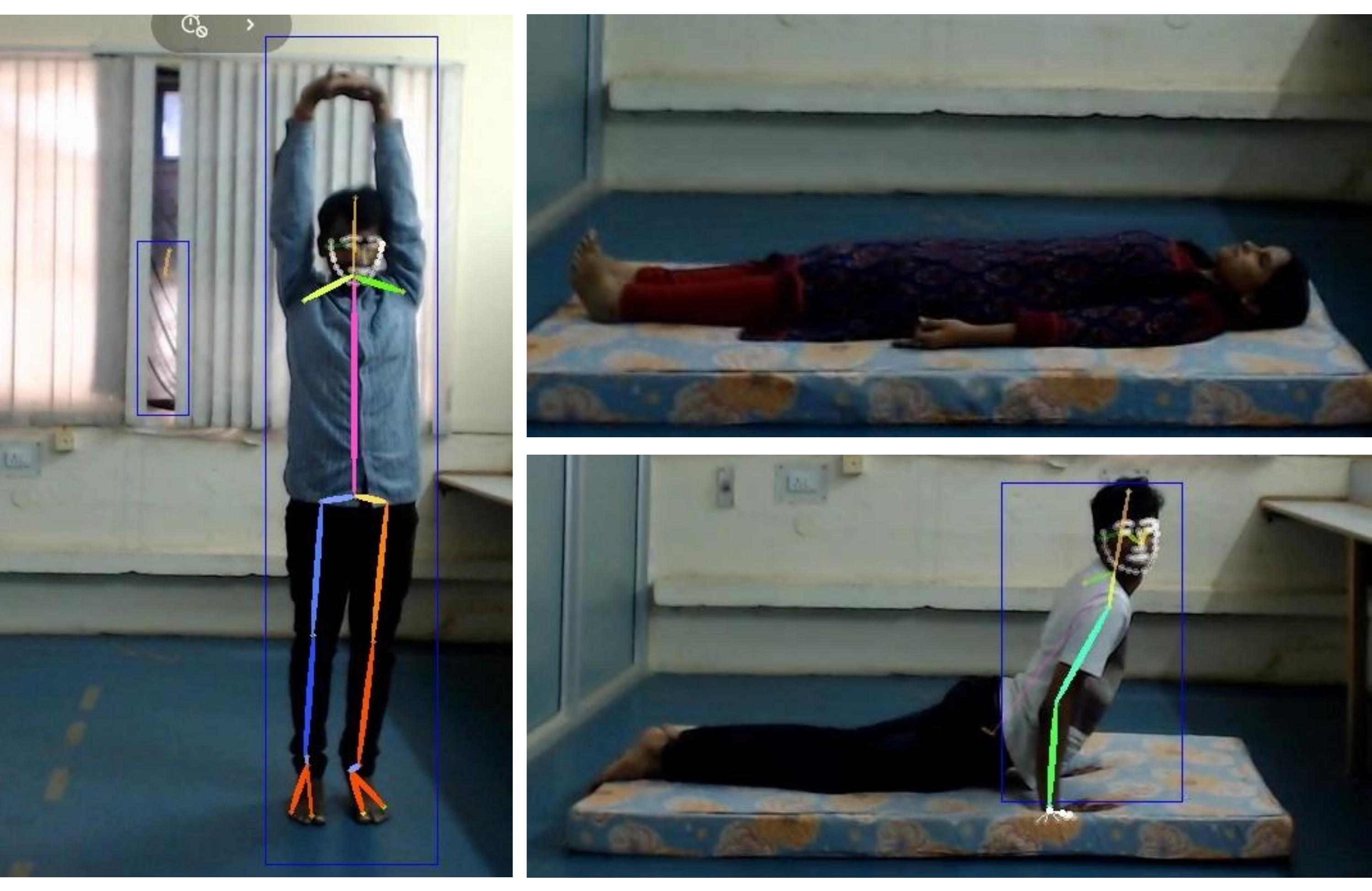}
         \caption{AlphaPose poor predictions. In left, hands are missed. In top right, complete human is missed, and a false positive detected. In bottom right, the bounding box itself is incorrect, affecting normalisation.}
     \end{subfigure}
     
    \caption{Some wrongly classified examples in Yoga-82 \cite{verma2020yoga} benchmark and Yadav et al \cite{yadav2019real} dataset.}
    \label{fig:wrongs}
    
    \vspace{-2mm}
    
\end{figure*}

The subject-wise results were, however, more acceptable. The exact results can be seen in Table~\ref{tab:yadav_subject}. These results further demonstrate the robustness of our evaluation strategy. Some examples where the model performed badly can be seem in Fig~\ref{fig:wrongs}. Note that in majority of the wrongly classified cases, the key-points detected by AlphaPose \cite{fang2017rmpe} are at fault. In some, there is a second human in sight, while the others are mislabelled. This comparatively poor performance of AlphaPose \cite{fang2017rmpe} on yogasana data points to a need for new key-point annotated datasets containing some of the difficult poses common here.

\subsection{Experiments on Jain et al. dataset}

The Jain et al.~\cite{jain2021three} dataset comprises of 27 subjects performing 10 different poses. The authors propose the use of 3D CNNs on video segments of 16 frames each. Nonetheless, we analyse the performance of our image based classification on their dataset. Again, we sample 100 frames from each video and use these frames for further analyses. Although this dataset consists of different subjects being recorded, we could not determine which subjects were involved in which video, and thus, will not use our subject-wise evaluation strategy here. These videos were recorded from a single uniform camera angle, and thus, we cannot perform our camera-wise evaluation either.

For frame-wise evaluation, we extracted a total of 23090 high resolution frames and re-scaled them to lower resolution because of resource constraints. The frame-wise results of our method on this dataset can be found in Table~\ref{tab:jain_frame}. Even with image-based prediction instead of video segments, our method outperformed their 3D CNN-based method that gave an accuracy of 91\%. This is a clear demonstration of the advantage of using transfer learning over end to end training in this data scarce domain.


\begin{table}[]
    \caption{Frame-wise results for Jain et al. \cite{jain2021three} dataset}
    \label{tab:jain_frame}
\begin{tabular}{clccc}
\toprule
ID & Yoga Pose & Precision & Recall & F1 Score \\
\midrule
0 & Garland Pose & 99.27\% & 98.27\% & 98.76\% \\
1 & Happy Baby Pose & 99.39\% & 98.65\% & 99.02\% \\
2 & Head To Knee Pose & 98.78\% & 98.59\% & 98.69\% \\
3 & Lunge Pose & 96.70\% & 96.83\% & 96.76\% \\
4 & Mountain Pose & 97.78\% & 99.10\% & 98.44\% \\
5 & Plank Pose & 97.88\% & 96.41\% & 97.14\% \\
6 & Raised Arms Pose & 94.02\% & 97.21\% & 95.59\% \\
7 & Seated Forward Bend & 98.63\% & 98.76\% & 98.69\% \\
8 & Staff Pose & 98.85\% & 98.47\% & 98.66\% \\
9 & Standing Forward Bend & 94.57\% & 93.83\% & 94.20\% \\
\midrule

 & Average (Ours) & \textbf{97.59\%} &	\textbf{97.61\%} & \textbf{97.59}\%\\
 & Average (Jain et al. \cite{jain2021three}) & 91\% &	91\% & 91\%\\
\bottomrule
\end{tabular}
\end{table}

\subsection{Experiments on Yoga-82 dataset}

Yoga-82 \cite{verma2020yoga} is one of the few openly available benchmarks for yoga classification. Although the authors have provided the links to download each image, some links were not reachable. We were able to extract 18k images out of the total 28k in the benchmark. We further curated the data to remove clipart-like images and finally had 12k images. We use this subset of 12k images for all our further analysis. The AlphaPose \cite{fang2017rmpe} key-points dataset for these images will be made openly available for reproducibility.

\begin{table}[]
    \caption{Accuracies for all levels of hierarchy in Yoga-82 \cite{verma2020yoga}. The performance of their three variants, MobileNet-V2 and DenseNet-201 are taken directly from the Yoga-82 paper.}
    \label{tab:yoga_82}
    
\begin{tabular}{c|ccc}
\toprule
     & \multicolumn{3}{c}{Top-1 Accuracy}\\
Method  & Level 1 & Level 2 & Level 3\\
\midrule
MobileNet-V2 \cite{sandler2018mobilenetv2} &  - & - & 71.11\%\\
DenseNet-201 \cite{huang2017densely} &  - & - & 74.91\%\\
Yoga-82 Variant 1 \cite{verma2020yoga} & 83.84\% & 85.10\% & 79.35\%\\
Yoga-82 Variant 2 \cite{verma2020yoga} & 89.81\% & 84.59\% & 79.08\% \\
Yoga-82 Variant 3 \cite{verma2020yoga} & 87.20\% & 84.42\% & 78.88\% \\
\midrule
Ours (DCPose\cite{dcpose})$^*$ & 86.47\% & 85.05\% & 55.07\%\\
Ours (KAPAO\cite{kapao})$^*$ & 86.53\% & 83.75\% & 78.01\%\\
Ours$^*$ & \textbf{91.21\%} & \textbf{87.91\%} & \textbf{80.14\%}\\
\bottomrule
\end{tabular}

\vspace{2mm}

\caption*{$^*$Our analysis is on a subset of original dataset. So, the performance is only representative, not for exact comparison.}

\vspace{-4mm}

\end{table}

To better demonstrate the benefits of using transfer learning, we have kept the classifier stage as simple as possible. For all of the analysis above, we used a simple random forest classifier. 
However, random forests were performing quite poorly on the more challenging Yoga-82 \cite{verma2020yoga} dataset. Using an \textit{ensemble} of Random Forests~\citep{breiman2001random}, Gradient Boosting~\citep{friedman2002stochasticGB} and LightGBM~\citep{ke2017lightgbm}, we obtained results comparable to those reported by the authors. A brief description of some wrongly classified samples can be seen in Fig~\ref{fig:wrongs}.

We also compare the method with the recently proposed methods DCPose\cite{dcpose} and KAPAO\cite{kapao}, but observe somewhat poorer results. We believe this may be because of lesser number of keypoints (17 over the body) and absence of temporal information required for DCPose. The key-points inferred by all these methods for Yoga-82 images will be made openly available for reproducibility. Our results along with those obtained by the Yoga-82 \cite{verma2020yoga} authors and some baselines used by them can be found in Table~\ref{tab:yoga_82}. 

In contrast to our dataset and Yadav et al \cite{yadav2019real} dataset, Yoga-82 consists of images in the wild. Along with a wider variety of subjects and poses, their images are also from many different camera angles. In addition to this, the images are not in a controlled environment, and background conditions are not uniform. All these factors make this a more challenging benchmark, and thus, our model has lower performance here compared to the other datasets. Other models in Table~\ref{tab:yoga_82} are deep architectures with a high level of complexity. Leveraging transfer learning, we were able to outperform them by training relatively simpler classifiers. This is clear validation of the usefulness of transfer learning in this data-scarce field of yoga classification. 


\section{Discussion}

While in our experiments we use AlphaPose\cite{fang2017rmpe}, a similar approach can be applied using other, more recent pose estimation methods. We experiment with KAPAO\cite{kapao} and DCPose\cite{dcpose} on Yoga-82 to observe the effect of various pose estimation methods on the first stage of our pipeline. To demonstrate the benefit of transfer learning on this data-scarce domain, we show that even with a simple classifier, the keypoints learnt by AlphaPose lead to quite promising results. Consistent performance on the 4 considered datasets demonstrates the efficacy of this approach. While previous Yogasana classification methods employed complex algorithms with end-to-end training, our simpler method is able to achieve comparable results. 

Out of the three evaluation strategies we explore, frame-wise evaluation is by far the most commonly used. Most existing work in this domain use this strategy itself. However, when using frames that have been extracted from videos, this strategy leads to the serious problem of target leakage. Yadav et al. \cite{yadav2019real} used this strategy and reported 100\% accuracies in half of the asanas they experimented with. Our results in Table~\ref{tab:frame_wise} are also extremely high, with 100\% precision in two poses. A similar trend can be seen with Jain et al \cite{jain2021three} dataset. Because of target leakage, we believe that frame-wise evaluation is not an ideal testing strategy.

With subject wise evaluations, our model gets tested on subjects it has never seen during training. We obtain good results with this strategy as well, as can be seen in Table~\ref{tab:subj_wise}. This suggests that our model is robust over variations in the subject performing the asana. This is a direct influence of using Transfer Learning in our first stage pose estimator. Although the classifier has not been trained on a particular subject, the pose estimator has been trained on a much wider variety of humans, albeit from a different dataset. This allows it to be robust over variations in the subjects and since the classifier only requires key-points from this pose estimator, this good performance is directly propagated to the final results. Similarly other desirable properties of a pose estimator would be propagated to the final results and so, advances in the field of yogasana pose classification also encourages development and research in the field of pose estimation.

The third evaluation strategy we use is camera-wise evaluation. In Table~\ref{tab:cam_wise}, it can be seen that as the model is trained on lesser number of camera angles, it tends to perform worse. This suggests that inclusion of more camera angles during training has a positive impact on the model's performance and generalizability. Even for the same pose being considered, different camera angles would be producing very different key-point coordinates. We believe that including all these different coordinates should allow a model to learn better quality view independent models, thus improving its performance. 

The results of using KAPAO\cite{kapao} and DCPose\cite{dcpose} on Yoga-82 shows no gain in performance over our experiments with AlphaPose\cite{fang2017rmpe}, though these two methods outperform AlphaPose significantly on standard datasets. We believe that this is because the model of AlphaPose that we use is pre-trained on the Halpe Full Body \cite{li2020pastanet} dataset which has 136 keypoints annotated on the human body while the model of DCPose is pre-trained on the PoseTrack \cite{posetrack} dataset which contains 17 keypoints and KAPAO is pre-trained on the COCO\cite{lin2014microsoft} dataset which consists of 17 keypoints. Since Yoga poses involve the human body to take poses which are extremely non-linear, more keypoints on the body are required to successfully classify a Yogasana. Moreover, the performance of DCPose is lower possibly because DCPose is primarily a pose-tracking model suitable for video data, requiring the previous and the next frame of the current frame to improve the pose estimation of the current frame. Since Yoga-82 is not a video dataset, there is no temporal continuity from one image to the next which can possibly degrade the performance of DCPose on the dataset which then results in poor performance of the classifier.

Besides the model itself being robust to camera angle variation, a different direction could be direct learning of a view independent representation of pose. Ongoing research on 3D pose estimation using only video data provides a new direction where detected poses from very different camera angles can have identical pose representations, thereby increasing generalizability over camera angles. Although particularly prominent in Yogasana postures, the need to generalize across different camera angles is also evident in many other applications of human pose estimation. A particularly challenging scenario where many pose estimation algorithm fail is occlusion, where part of a body part is hidden or not directly visible. It is noteworthy to see that a body part occluded when viewed from one camera angle may actually be clearly visible from another. If a model is truly view independent, it should be able to get the pose from non-occluded view and use it for the pose in the occluded view. Thus, the study of view independent methods of pose classification may help mitigate the problem of occlusion. A pose classification model that generalizes to unseen camera angles would be quite beneficial to the community. 

Yoga-82 \cite{verma2020yoga} has images in the wild, with significant variation in the camera angles, subjects, background lighting conditions and many other factors. The performance on this dataset for the level 3 classes (82 asanas) is very similar to our camera-wise evaluations, where variations other than that of the camera angle are minimal. This suggests that the camera angle variations have the most impact on the performance of a method, compared to frame wise and subject wise variations. Thus, a view independent classification framework that can effectively evaluate the generalizability of the model to different camera angles would provide a much better estimate of the performance of the model for a dataset in the wild. We hope this work can serve as a strong baseline for our and other researchers' subsequent works.


\section{Conclusion and Future Work}
\label{section:conclusion}

In this paper, we propose a two-stage architecture for classification of yoga poses. We use AlphaPose \cite{fang2017rmpe}, a well established pose estimation method as the feature extractor in the first stage and a random forest classifier in the second stage. In the absence of large scale datasets for yoga pose estimation, we use models pre-trained on existing large scale dataset Halpe Full Body \cite{li2020pastanet}. Transfer learning from pose estimation models is expected to result in better performance for Yogasana classification. We create a new dataset for Yoga Pose classification that focuses on different views of a subject. To the best of our knowledge, ours is the first yoga dataset explicitly considering asanas from 4 different camera angles, and has the largest number of subjects being recorded systematically. The key-points data as well as model codes used will be made open source to further accelerate research in this area. 

We devise a 3 step evaluation scheme to evaluate a model's robustness to different sources of variation. In light of particular shortcomings with the individual evaluation methods, we advocate the use of all three to effectively test generalizability of a model. We demonstrate that our model performs admirably on the first two parts and is competitive with previously published Yogasana classification methods on 3 publicly available datasets. Observing a noticeably lower performance when evaluated across camera angles, we discuss the particular challenges involved in this setting and argue that this camera wise evaluation can often be more reliable as a metric. 

There are a few relevant limitations to our approach. Since we rely on pose estimation and accurate detection of keypoints in our first stage, most challenges present in pose estimation naturally reflect in our approach as well. Occlusion and inversion are two commonly studied failure cases. These cases are particularly challenging in the case of Yogasanas because of the complex poses involved here, where some body parts can often be difficult to distinguish or hidden from view because of other body parts. Most large scale pose estimation datasets include normal day to day postures that may not allow a model to perform well on such complex postures. Even in the cases where occlusion occurs, hidden body parts are usually not labelled. Our dataset also contains classification labels only and significantly higher effort and time would be required to collect accurate keypoint annotations for it. Besides scarcity of data, poor generalizability to camera angles is also a limitation for our approach. We believe future work can strive to alleviate many of these limitations. The limitations show significant scope for further research in view independent pose classification and a need for large scale pose estimation datasets with complex poses similar to Yogasanas. 3D pose estimation methods and use of depth information to label occluded or hidden body parts may also help mitigate some of these challenges.


In applications such as that of an automated Yoga trainer, it will be desirable to have models that are robust across different camera angles. To achieve this, either one needs to train models on larger data employing many different camera angles (our results indicate incorporating more cameras in training set improves the test performance), or one needs to incorporate computer vision methods to make these models independent of camera angle. Our results on unseen camera angles show that there is still significant scope for improvement in classification of Yogasanas from different camera angles. This is an exciting area for future study. We believe that subsequent research in camera view independent Yogasana classification will also find many other applications. It is also likely to advance the state-of-the-art in the area of human pose estimation.


\bibliographystyle{ACM-Reference-Format}
\bibliography{references}












\begin{table*}[]
    \caption{Description of our dataset}
    \label{tab:describe_data}
\begin{tabular}{cl|cccc|ccccc}
\toprule
& & 
\multicolumn{4}{|c}{Number of Subjects} &
\multicolumn{5}{|c}{Number of Frames}  \\
ID & Yoga Pose & Camera 1 & Camera 2 & Camera 3 & Camera 4 & Camera 1 & Camera 2 & Camera 3 & Camera 4 & Total\\

\midrule
0 & Garudasana left & 35 & 33 & 32 & 12 & 1916 & 1738 & 1724 & 622 & 6000\\
1 & Garudasana right & 40 & 35 & 33 & 9 & 2129 & 1807 & 1613 & 451& 6000 \\
2 & Gorakshasana & 36 & 33 & 32 & 7 & 2040 & 1852 & 1706 & 402 & 6000\\
3 & Katichakrasana& 47 & 43 & 39 & 15 & 1962 & 1798 & 1592 & 648& 6000 \\
4 & Natavarasana left & 37 & 26 & 27 & 5 & 2369 & 1651 & 1677 & 303& 6000\\
5 & Natavarasana right  & 32 & 25 & 26 & 1 & 2316 & 1792 & 1826 & 66 & 6000\\
6 & Pranamasana left & 36 & 32 & 29 & 4 & 2148 & 1879 & 1739 & 234 & 6000\\
7 & Pranamasana right & 34 & 31 & 26 & 2 & 2143 & 2026 & 1710 & 121& 6000\\
8 & Tadasana & 41 & 36 & 38 & 18 & 1807 & 1596 & 1746 & 851 & 6000\\
9 & Vrikshasana left & 46 & 42 & 42 & 11 & 1913 & 1787 & 1791 & 509 & 6000\\
10 & Vrikshasana right & 38 & 32 & 32 & 4 & 2211 & 1789 & 1784 & 216 & 6000\\
11 & Still & 36 & 31 & 27 & 2 & 2231 & 1972 & 1693 & 104& 6000 \\	
\midrule
\multicolumn{2}{c|}{Total} & - & - & - & - & 25185 & 21687 & 20601 & 4527 & 72000 \\
\bottomrule
\end{tabular}
\end{table*}

\begin{figure*}[!htb]
     \centering
     \begin{subfigure}[b]{0.23\textwidth}
         \centering
         \includegraphics[width=\textwidth]{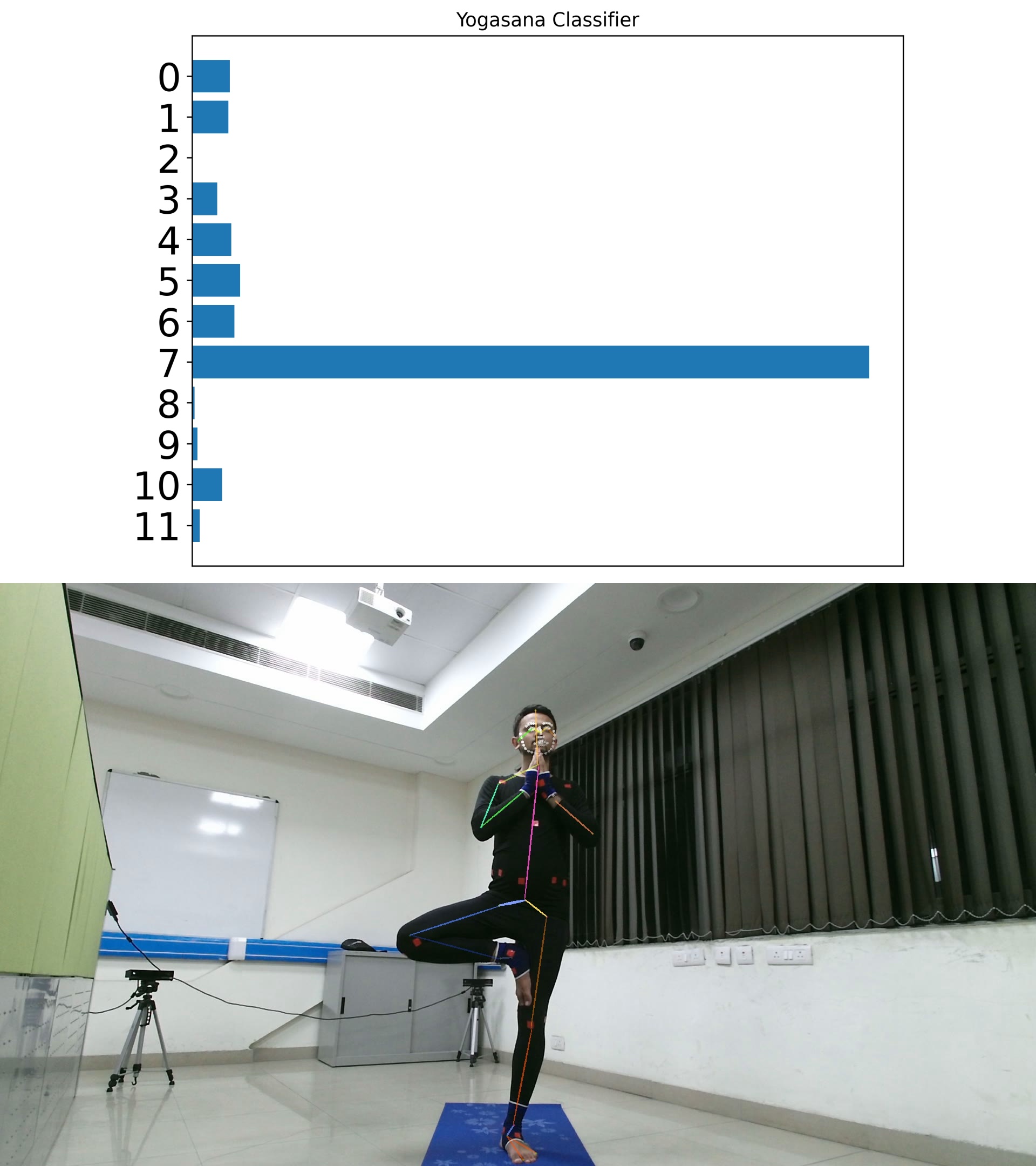}
     \end{subfigure}
    \hfill
     \begin{subfigure}[b]{0.23\textwidth}
         \centering
         \includegraphics[width=\textwidth]{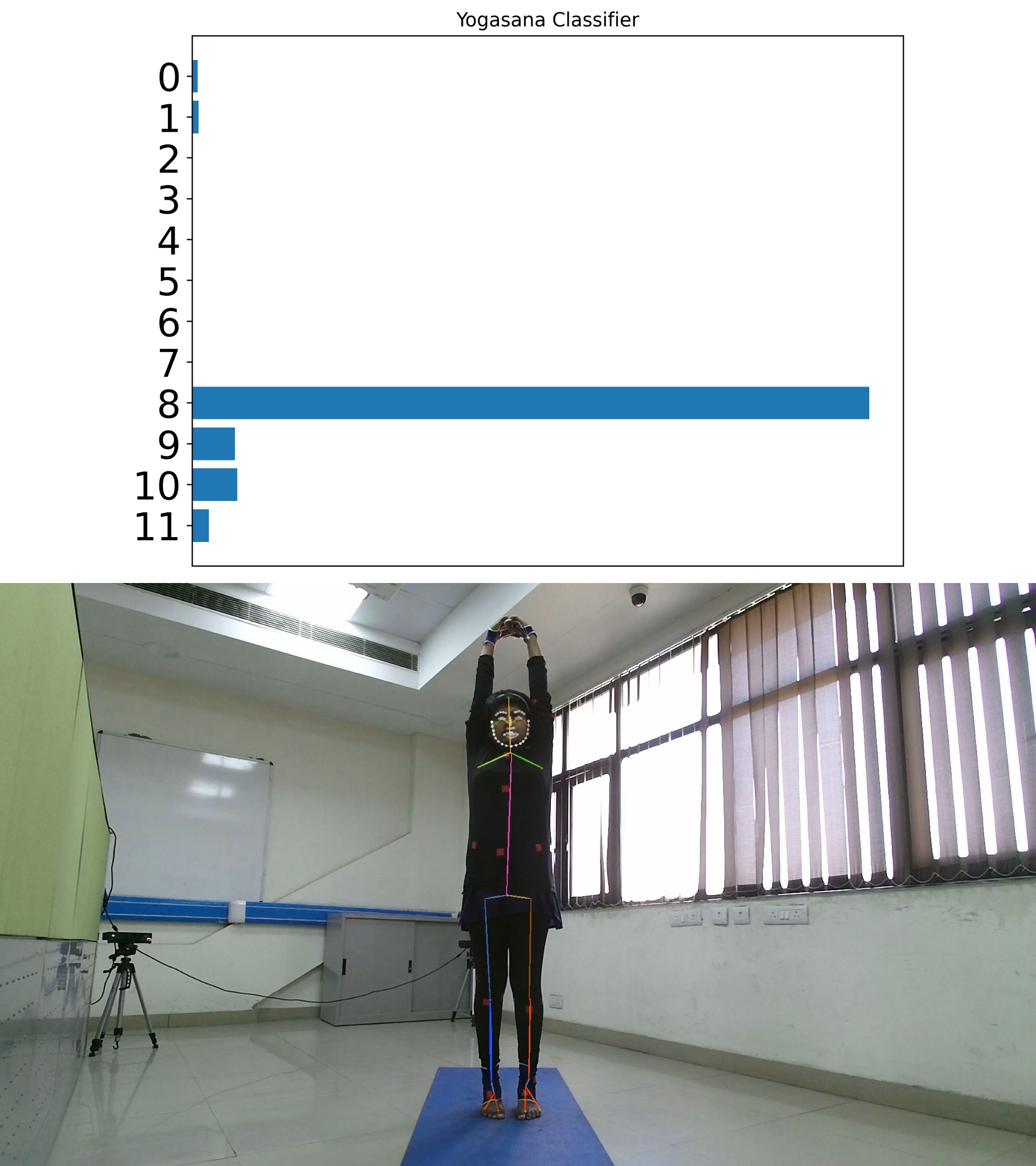}
     \end{subfigure}
    \hfill
     \begin{subfigure}[b]{0.23\textwidth}
         \centering
         \includegraphics[width=\textwidth]{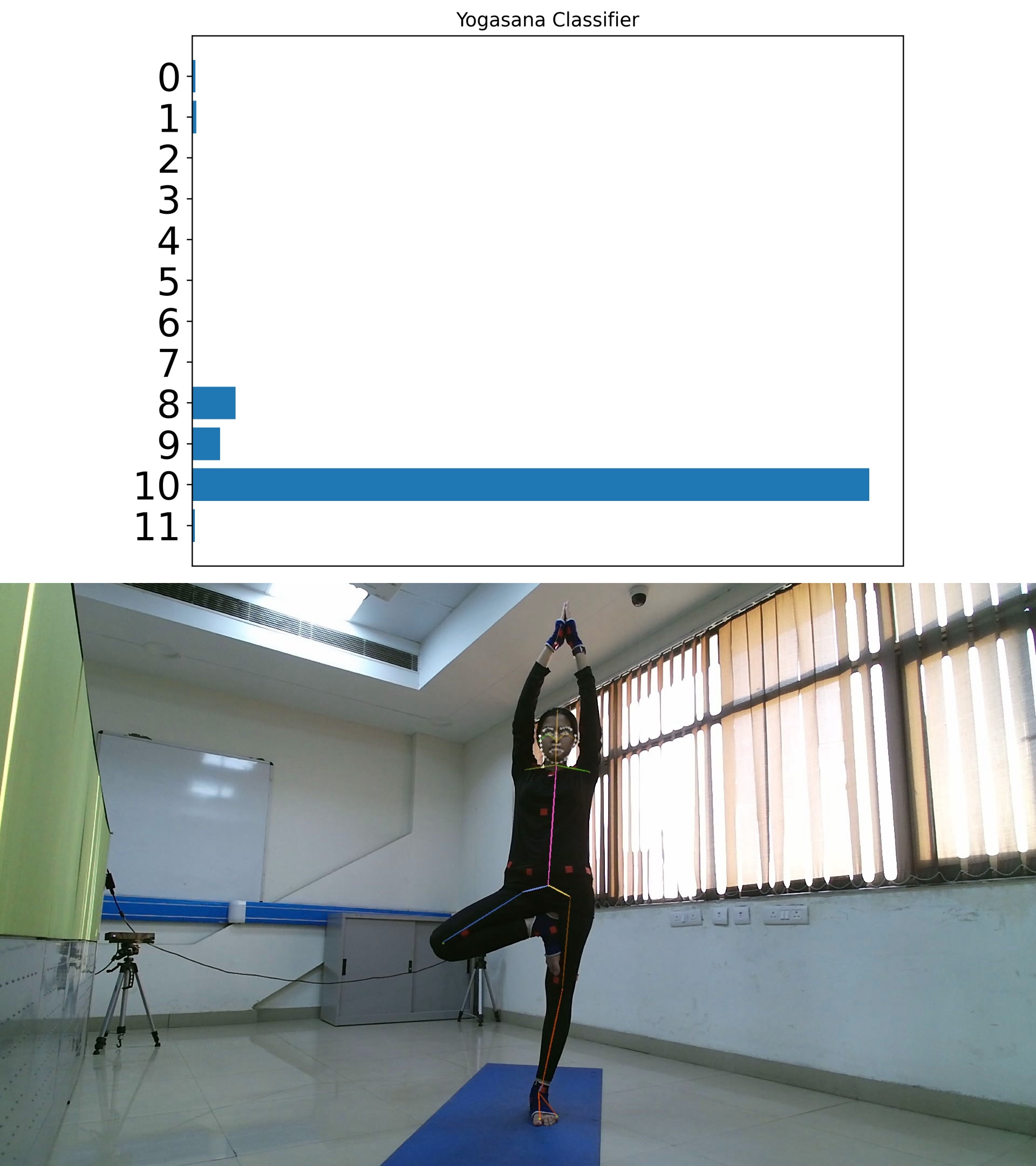}
     \end{subfigure}
    \hfill
     \begin{subfigure}[b]{0.23\textwidth}
         \centering
         \includegraphics[width=\textwidth]{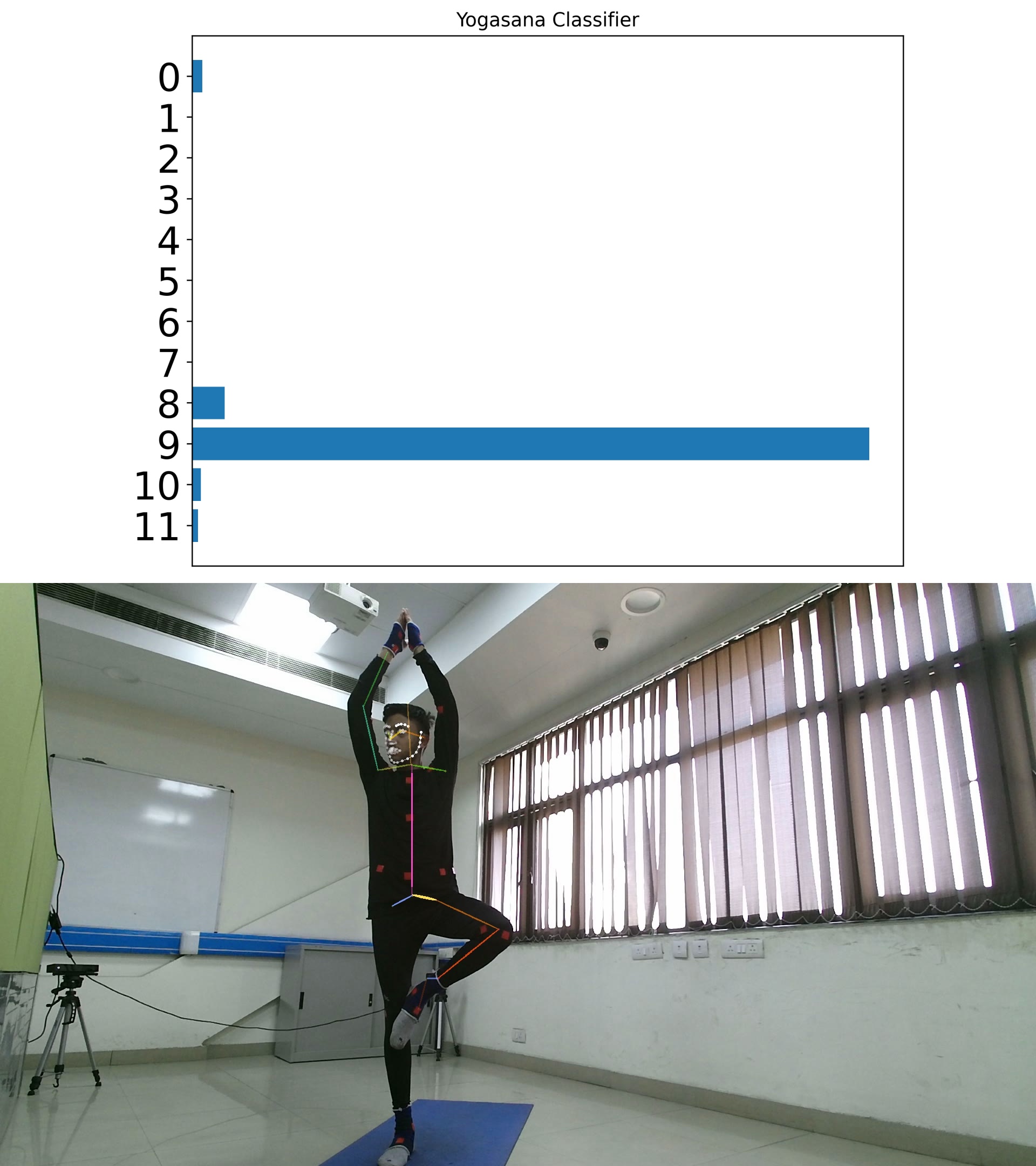}
     \end{subfigure}
     \caption{Some samples with good predictions. The model is able to distinguish between left and right for bilateral poses}
     \label{fig:samples}
\end{figure*}

\begin{figure*}[!htb]
     \centering
     \begin{subfigure}[b]{0.23\textwidth}
         \centering
         \includegraphics[width=\textwidth]{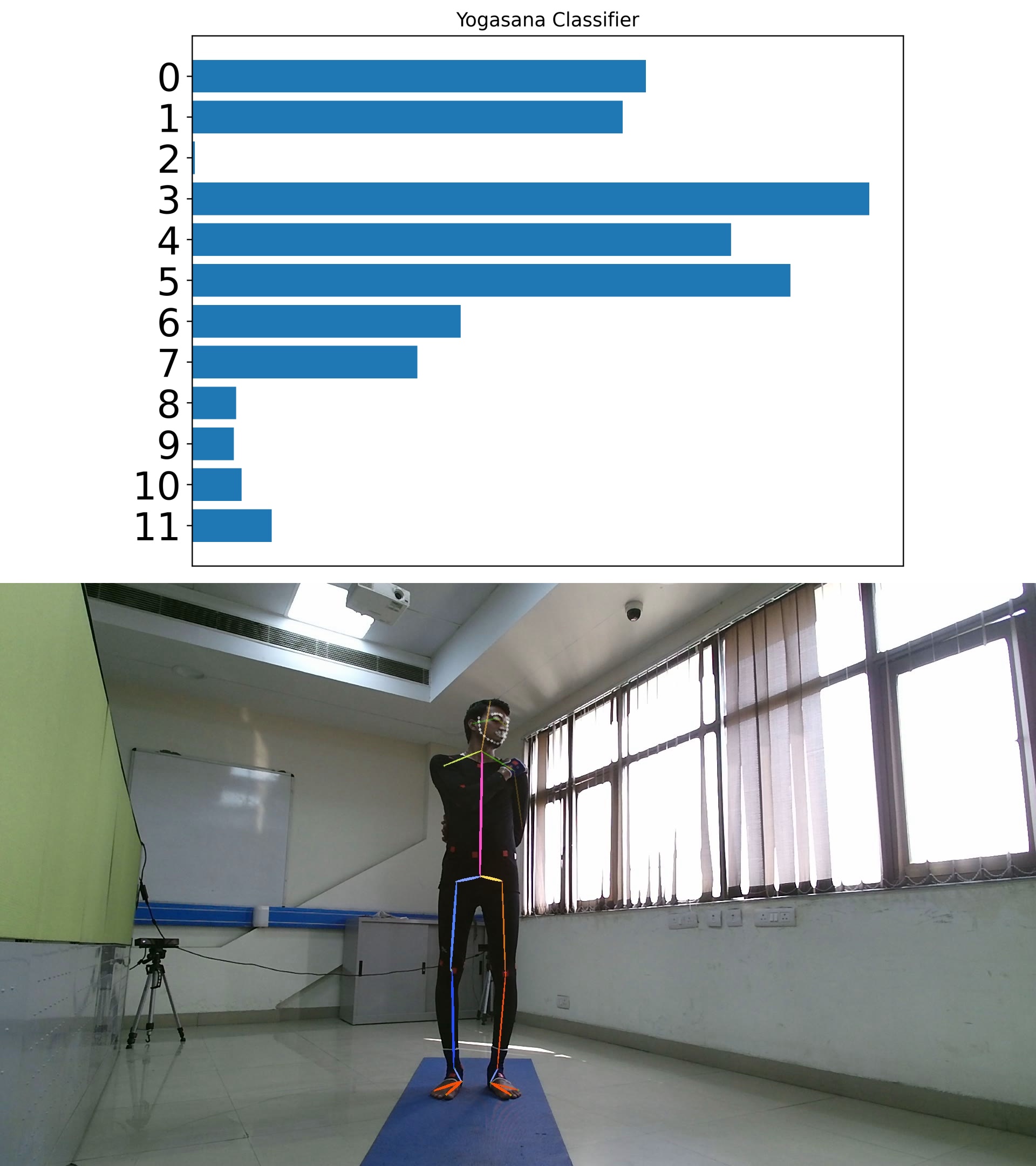}
     \end{subfigure}
     \hfill
     \begin{subfigure}[b]{0.23\textwidth}
         \centering
         \includegraphics[width=\textwidth]{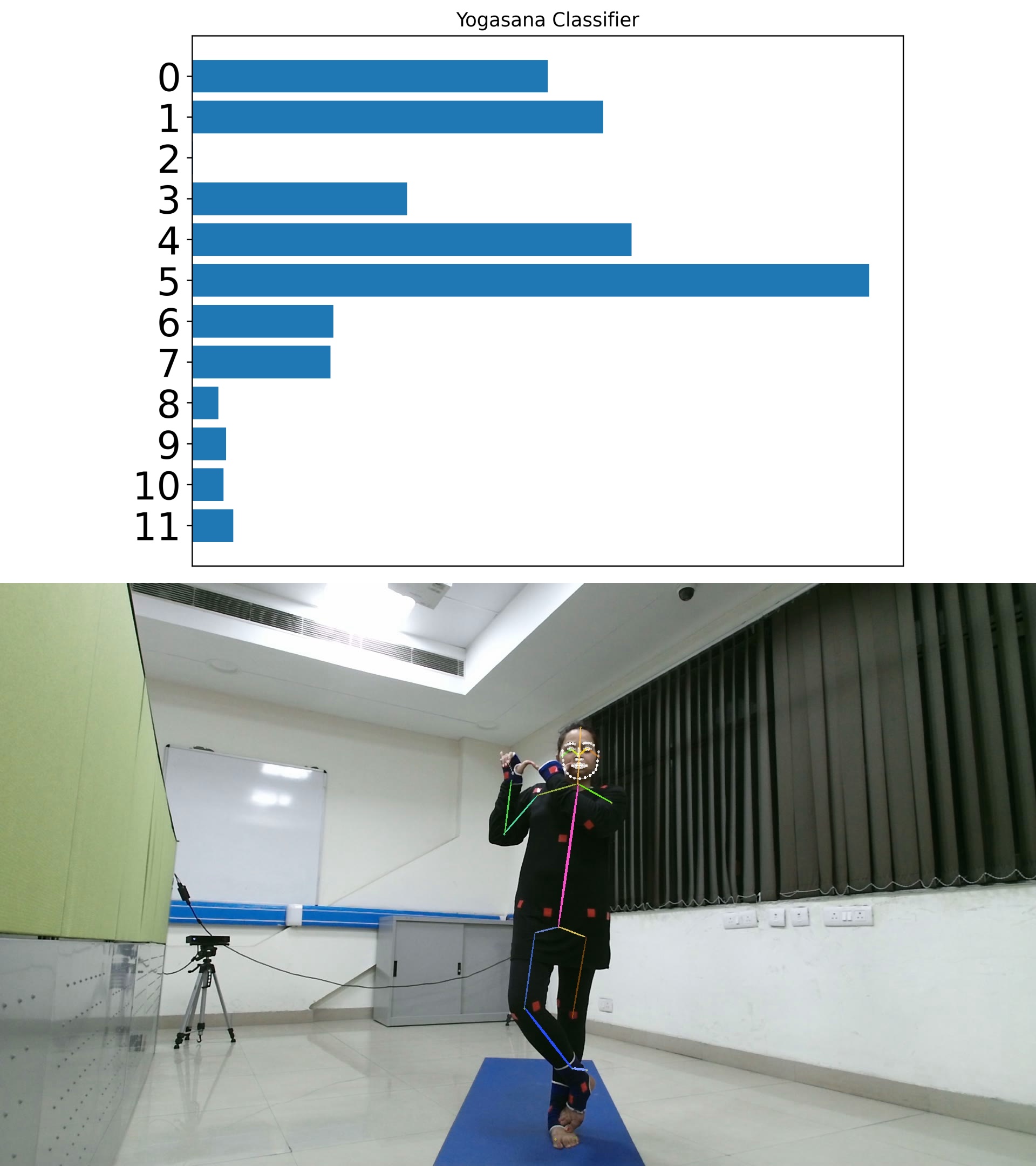}
     \end{subfigure}
    \hfill
    \begin{subfigure}[b]{0.23\textwidth}
         \centering
         \includegraphics[width=\textwidth]{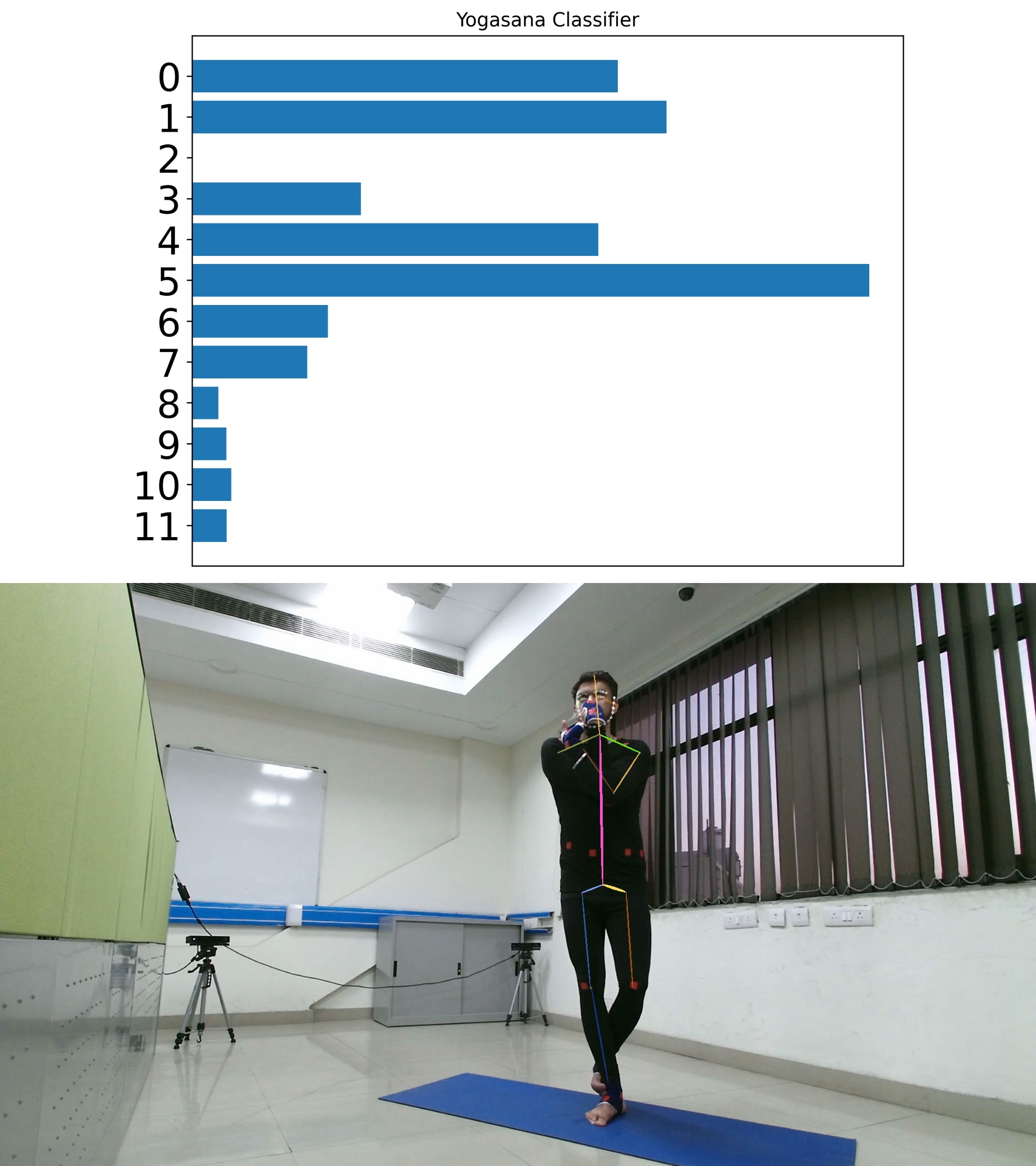}
     \end{subfigure}
     \hfill
     \begin{subfigure}[b]{0.23\textwidth}
         \centering
         \includegraphics[width=\textwidth]{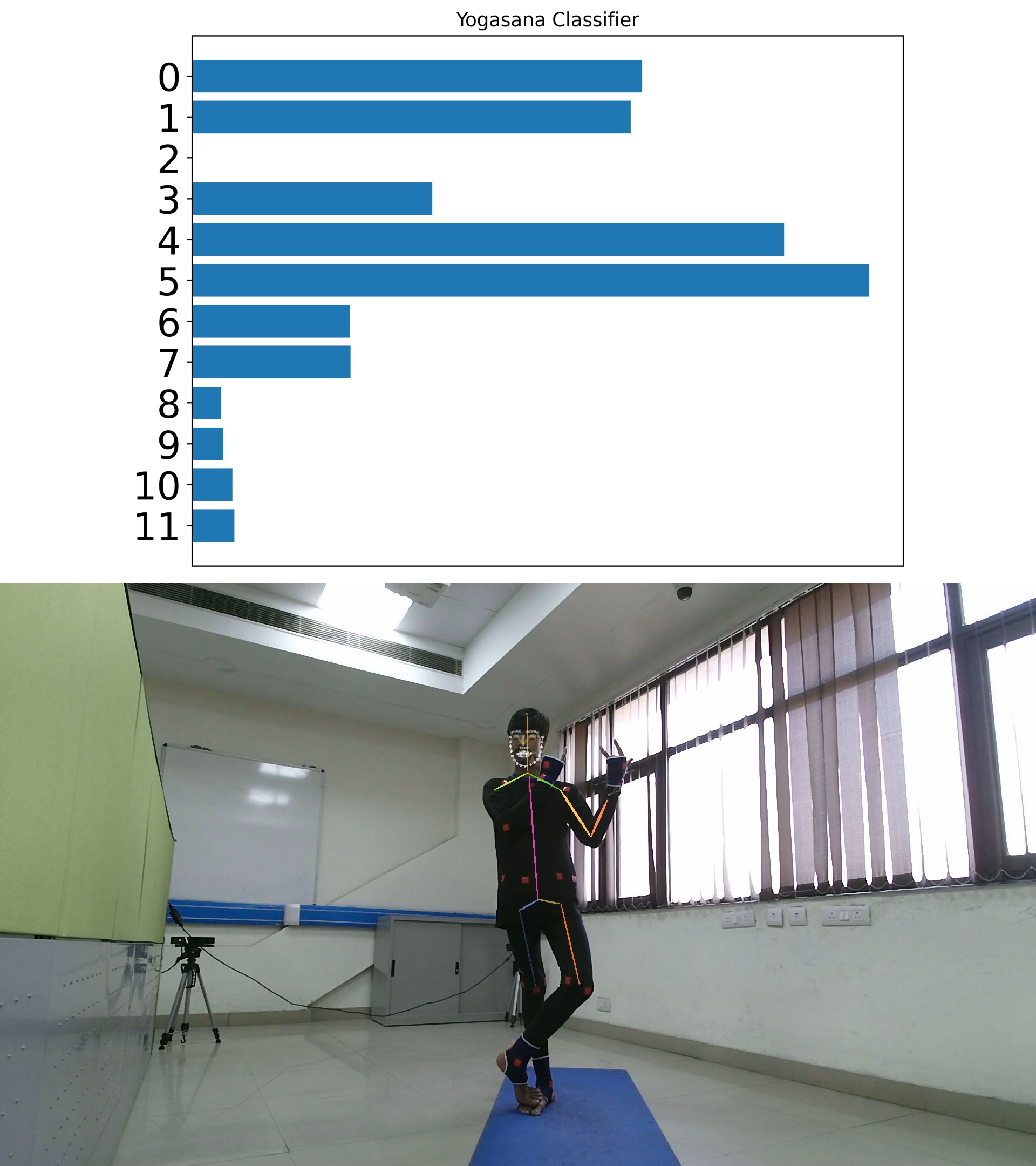}
     \end{subfigure}
   
     \caption{Some samples with poor predictions. The poses are highly symmetric and the classifier has quite a bit of confusion.}
     \label{fig:samples2}
     
     \vspace{-4mm}

\end{figure*}


\section*{Supplementary Material}

\appendix

\section{Dataset Description}

\subsection{Recording Yoga Videos}

Videos of subjects performing different yoga poses were recorded in a systematic way. We had 51 subjects performing 20 asanas, and they were being recorded from 4 different camera angles. Due to some constraints, some of the asanas could not be performed by some subjects and some videos could not be recorded by some camera angles. All in all, a total of 3532 videos were recorded. Along with the raw videos in avi format, the IR and depth data was also collected using Microsoft Kinect. In addition to these, Kinect itself also provided keypoints detected by it, and these keypoints were also saved for future use.

The videos were recorded in a controlled indoor environment with sufficient lighting. This allowed us to precisely monitor the effect of each component of the dataset. For example, while analysing the performance of our model with unseen subjects, the camera angles were comparatively more evenly distributed, and while analysing the same with unseen camera angles, the subjects were distributed in both. This allowed us to decouple the effects of subject variation and camera angle variation. In contrast, a dataset recorded in the wild may have a larger variation, but would lack control. For example, if the performance on two subsets were to differ, it would be very difficult to pinpoint the exact cause of the difference. The difference could be because of the subject, the camera angle the lighting conditions or even the background.

\subsection{Running AlphaPose}

In the envisioned Yoga Pose classifier, recorded videos would be processed offline, while for real-time working, it would be a part of the complete pipeline. For our recorded videos dataset, we ran AlphaPose inference on each videos and stored them separately. The open-source AlphaPose model gives as output a json containing the key-points, their scores, and the corresponding human's detected bounding box for each detection in each frame of each video. In addition to the key-point coordinates, it also provides visualisations to qualitatively view the results. Once the AlphaPose predictions were obtained for all the videos, we moved on to the next stage of frame extraction.

\begin{figure*}[!htb]
     \centering
     \begin{subfigure}[b]{0.48\textwidth}
         \centering
         \includegraphics[width=\textwidth]{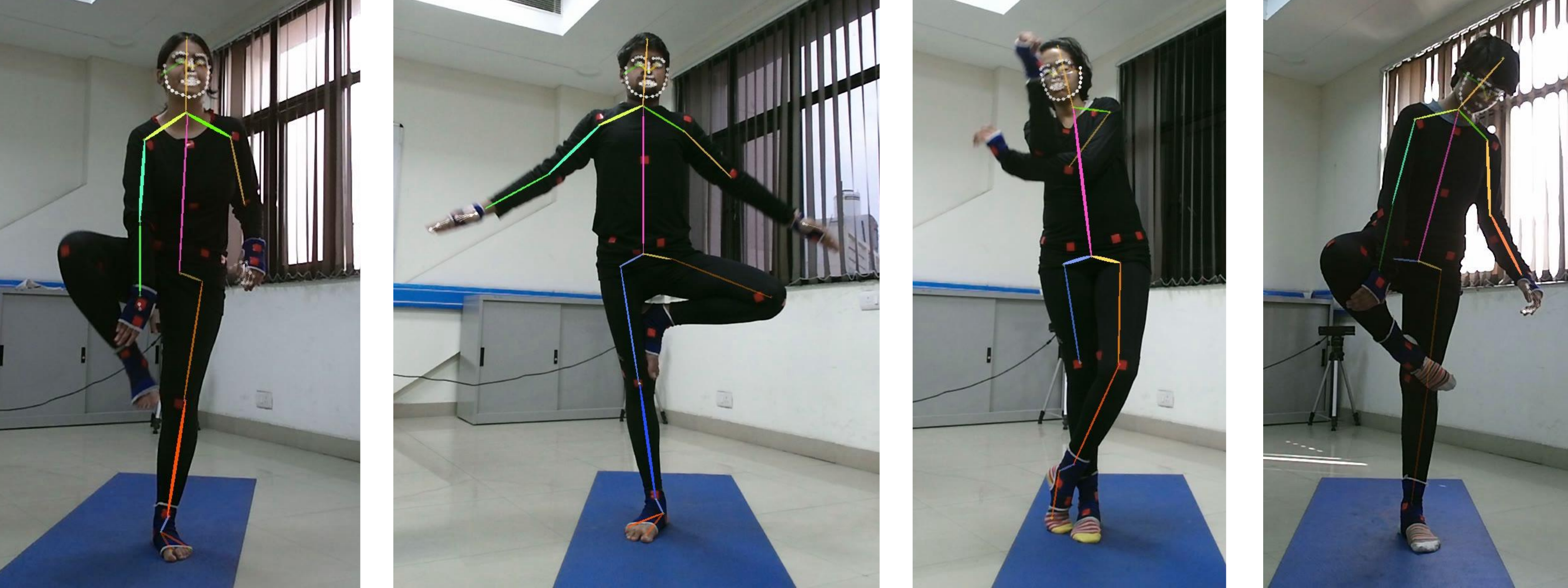}
         \caption{Some examples with subjects in transition between asanas}
         \label{fig:transition}
     \end{subfigure}
     \hfill
     \begin{subfigure}[b]{0.48\textwidth}
         \centering
         \includegraphics[width=\textwidth]{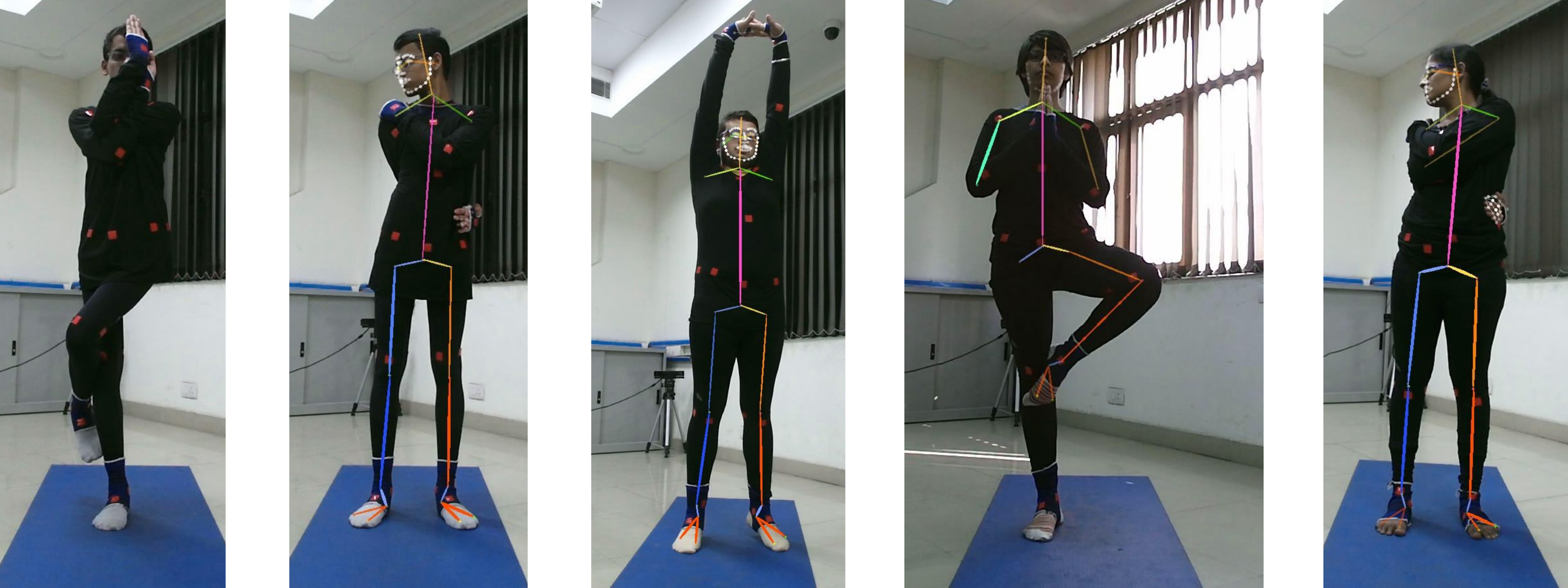}
         \caption{Some examples with poor AlphaPose predictions.}
         \label{fig:wrong}
     \end{subfigure}
    
     \caption{Some examples of mis-classified frames.}

\end{figure*}

\subsection{Frame Extraction}

We planned to work on an image-based classifier instead of video-based in order to achieve near real-time performance with the complete pipeline. Thus, our classifier needed image-based data to train and test. We obtain this data by extracting individual frames from the recorded videos and using their corresponding detected key-points. Our videos being around 2-5 minute each and being recorded at $30fps$, the total number of frames was too large for our purposes. Instead of taking all the frame in a video, we uniformly sampled roughly 200 frames in each part of the video. Again, since subjects would be performing both left and right poses for bilateral poses, we treated each of them separately, and extracted 200 frames for both left and right poses. 

We had recorded the timestamps corresponding to each part of the pose. We observed that the timestamps between cameras were slightly different, and thus there was a possibility of mis-labelling. Also, in the videos, there was usually a \textit{transition} period, where either the subject is getting into the pose, changing from left to right, or going out of it. We took a buffer of 1 second before and after each timestamp for these transition periods and sampled our frames from the remaining section only. 

Aside from these, we observed that in some instances, the subject temporarily loses balance and is not in the actual pose for some time. We were not able to record the timestamps corresponding to each of these instances, and thus, some frames may be mislabelled because of it.

After extracting the frames from all the videos, we further subsampled the dataset as discussed in Section 3.1 of the full text. The distribution of frames across subjects and camera angles for each subsampled class can be seen in Table~\ref{tab:describe_data}. The distribution is a little skewed with respect to camera 4, and this was reflected in the performance of camera-wise evaluation, where including the smaller camera 4 and testing on the similarly-oriented larger camera 3 led to poor results. 

To the best of our knowledge, ours is the first yoga dataset explicitly considering asanas from 4 different camera angles, and has the largest number of subjects being recorded. Yoga-82, an openly available dataset, does have images that seem to have been taken from different camera angles. However, these images are in the wild, and the effects of camera angle variation cannot be isolated from other variations. We believe that decoupling the camera angle variation would allow us to better focus on generalising over it, and hope our dataset would help other researchers improve their methods as well.


\section{Qualitative Analysis}

In addition to the Quantitative evaluation in the full text, we also try to qualitatively evaluate the performance of our method on our dataset. Visualisations of the classification predictions for some samples can be seen in Fig~\ref{fig:samples} and Fig~\ref{fig:samples2}. Instead of displaying only the final predicted value, we visualise the probabilities given by the model for each class, or the `soft' labels. The probabilities are visualised as bar plots, with the y-axis having the asana id, consistent with the convention in Table~\ref{tab:describe_data}, and the x-axis the predictions for each class.

In Fig~\ref{fig:samples}, we present some good performing examples. In these examples, the predicted class is correct, and also, the probabilities for the other classes is comparatively low, meaning that the classifier has less confusion. Sub-figure*s 3 and 4 here are actually the left and right versions of the same yoga pose, and the classifier is able to differentiate between them easily as well. These examples consist of asanas that have some characteristics which makes them easier to differentiate, and thus, are easy for the classifier to predict correctly.

On the other hand, the examples in Fig~\ref{fig:samples2} have much more confusion. Although the predictions are actually correct, the classifier is also predicting high probabilities for wrong classes. Thus, the prediction may be completely wrong for another similar example.

\section{Mis-classified examples}

In this section, we discuss some examples where the camera-wise evaluation predicted wrongly. In Fig~\ref{fig:transition}, the subject is either yet to get into the main pose of the asana, has completed the asana and is coming out of it, or has lost balance in between. We were not able to accurately identify all these examples, and thus, approximately 2-3\% frames in our dataset may be mis-labelled. In our subsequent research, we plan to curate our dataset better by identifying all these mis-classified frames and removing them from the dataset. 

In Fig~\ref{fig:transition}, from left to right, a subject is getting into the main pose for Vrikshasana, a subject is changing from left to right for Vriskshasana, a subject loses balance while performing Garudasana, and a subject loses balance while performing Vriskshasana.

In Fig~\ref{fig:wrong}, we show some examples where AlphaPose performed poorly. The poses common in Yoga are not present in most existing datasets for pose estimation and HAR, and thus, the AlphaPose model is trained on comparatively easier images. This points to a need for pose estimation datasets having more complex poses.

In Fig~\ref{fig:wrong}, from left to right, no key-points are detected by AlphaPose for Katichakrasana, the key-points of arms and hands are missed for Katichakrasana, the arms are missed again for Tadasana, Vrikshasana and Katichakrasana.




\end{document}